\documentclass[10pt,twocolumn,letterpaper]{article}

\usepackage[datasets]{wacv}

\usepackage{graphicx}
\graphicspath{{assets/figures/}{figures/}}
\usepackage{tikz}
\usetikzlibrary{positioning,arrows.meta,shapes.geometric,fit,backgrounds,calc,patterns}
\usepackage{amsmath,amssymb}
\usepackage{amsthm}
\newtheorem{theorem}{Theorem}
\usepackage{booktabs}
\usepackage{multirow}
\usepackage{xspace}
\usepackage{enumitem}
\usepackage{url}

\newcommand{\invb}{\textsc{invB}\xspace}
\newcommand{\lofo}{\textsc{lofo}\xspace}
\newcommand{\lodo}{\textsc{lodo}\xspace}
\newcommand{\lomo}{\textsc{lomo}\xspace}
\newcommand{\allwrong}{\textsc{all-wrong}\xspace}
\newcommand{\auc}{\textsc{auc}\xspace}
\newcommand{\aurc}{\textsc{aurc}\xspace}
\newcommand{\K}{K\xspace}

\newcommand{\fsBefore}[1]{}
\newcommand{\fsCapAbove}[1]{}
\newcommand{\fsCapBelow}[1]{}
\newcommand{\fsAfter}[1]{}

\newcommand{\coderepo}{\href{https://github.com/lukegcollins/WACV27-HFEFCDD}{\nolinkurl{github.com/lukegcollins/}\allowbreak\mbox{\nolinkurl{WACV27-HFEFCDD}}}}

\providecommand{\keywords}[1]{}

\definecolor{wacvblue}{rgb}{0.21,0.49,0.74}
\usepackage[pagebackref,breaklinks,colorlinks,allcolors=wacvblue]{hyperref}
\hypersetup{pdftitle={Honest Fusion Evaluation for Cross-Domain Deepfake Detection: Leakage Audits and Tail Robustness},pdfauthor={Luke Collins, Chang-Tsun Li, Duc Thanh Nguyen}}
\usepackage[capitalize]{cleveref}

\def\wacvPaperID{3232}
\def\confName{WACV}
\def\confYear{2027}

\title{Honest Fusion Evaluation for Cross-Domain Deepfake Detection:
Leakage Audits and Tail Robustness}

\author{Luke Collins\qquad Chang-Tsun Li\qquad Duc Thanh Nguyen\\
School of Information Technology, Deakin University, Waurn Ponds, Australia\\
{\tt\small \{l.collins, chang-tsun.li, duc.nguyen\}@deakin.edu.au}
}

\begin{document}
\maketitle

\begin{abstract}
Cross-domain video deepfake detection increasingly uses frozen foundation-model experts. Their outputs can be combined by calibrated score averaging, by a per-video router that picks one expert, or by a learned head over concatenated features. Routing and feature fusion, the expressive options, are often reported to beat plain averaging. Across a $22$-dataset frozen-expert testbed, we find those gains are largely artifacts of how cross-domain performance is measured. A simple corner theorem locates the gap routing aims to close: because \auc{} depends only on the ordering of scores, the best any score-space router could do is an oracle that reads the test label to pick each video's expert, a ceiling no realisable, label-blind router reaches in general. The theorem does not order real routers against a well-calibrated global average; the case against routing is empirical, and here no router we test beats that average. Feature fusion reads raw features, falls outside the theorem, and is audited empirically. The reported gains trace to leakage: holding out one dataset leaks siblings that share identities and reals, over-crediting routing, while holding out one manipulation keeps that dataset's reals in training, over-crediting feature fusion. A $2{\times}2$ over fold and evaluation granularity pins the $+2.6$pp feature-fusion ``win'' to the fold alone: $12/12$ significant under leaky folds versus $0/12$ under honest folds, which hold out whole dataset families; there both gains vanish. Fusion still pays off where the theorem is silent, in the tail of hard videos and families: as experts are added, the fraction of videos every expert misclassifies drops from $15.2\%$ to $2.3\%$, and global calibrated fusion (\invb) beats the best single expert on worst-family \auc{}, selective risk (\aurc), and low-false-positive recall, significantly and stably across protocols. Practitioners should calibrate and combine globally, hold out whole dataset families for evaluation, and measure success in the tail. For score-space routing under ranking metrics, honest evaluation and expert diversity matter more than combiner complexity.
\end{abstract}
 \section{Introduction}
\label{sec:02_introduction}

Manipulated videos leave heterogeneous traces: face swaps introduce spatial blending artifacts, reenactment and audio-driven animation disrupt natural temporal dynamics, generative decoders leave frequency-domain residues, and any forgery distorts the geometry a face-identity model has learned from real people. No single detector captures all of these cues, and detectors tuned to one manipulation family routinely collapse on the next~\cite{yan2023deepfakebench,cui2025forensics,yermakov2026deepfake}. A popular recipe therefore combines several \emph{frozen} foundation-model experts (spatial~\cite{simeoni2025dinov3}, temporal~\cite{assran2025v}, spectral~\cite{frank2020leveraging}, and face-specific~\cite{wang2025fsfm}) into a single cross-domain detector, since freezing the backbones sidesteps the cost and instability of joint fine-tuning. This leaves one design choice: how to combine the experts. Three mechanisms dominate the literature: calibrated \emph{score fusion} (average or weight the experts' probabilities)~\cite{wolpert1992stacked,lakshminarayanan2017simple}, per-video \emph{routing} (a mixture-of-experts gate that picks an expert per input)~\cite{jacobs1991adaptive,shazeer2017outrageously,kong2026moe}, and \emph{feature fusion} (a head over the concatenated frozen features). Routing and feature fusion are frequently reported to outperform plain score fusion under cross-dataset and cross-manipulation protocols.

We revisit that premise and reach the opposite conclusion: \emph{for honest cross-domain ranking, the combiner barely matters, and the apparent gains of the expressive combiners are largely evaluation artifacts}. We call an evaluation \emph{honest} when every test video comes from a dataset family held out of combiner fitting, so that no sibling dataset sharing its source identities or pristine reals remains in the combiner's training data (one deliberate, leakier exception: \cref{sec:05a_setup}). Expert fusion still helps where it matters for deployment, in the \emph{tail}: the videos every expert misclassifies, the worst-served family, and the low-false-positive operating points. Our claims concern only \emph{post hoc} fusion of frozen experts. The theorem bounds a router through its \emph{output}, a convex mixture of the expert scores, whatever its gate reads, so a gate that reads content instead of scores faces the same unrealisable \auc{} ceiling; we test both a score-derived and a content-conditioned gate, and neither beats a calibrated global average. End-to-end joint fine-tuning, which changes the experts themselves, is a different question that we do not address.

\paragraph{Routing's \auc{} ceiling is an unrealisable oracle.}
We prove a \emph{corner theorem}~(\cref{thm:corner}): because $\auc$ is an ordinal functional of the combined score~\cite{cortes2003auc,menon2016bipartite}, the $\auc$-optimal rule inside the experts' score interval is a hard, label-aware router, and no soft router can exceed that corner under $\auc$. Because that optimum needs the test label to choose each video's expert, it is an \emph{unrealisable} ceiling that label-blind combiners and routers do not reach in general. The theorem thus locates the gap; it does not order \emph{realisable} routers against a calibrated global combiner. The case against routing is empirical: the per-video oracle winner is predictable out-of-family ($q = 0.84$), yet hard-routing it still \emph{loses} to \invb{}, and the soft routers we test do no better~(\cref{sec:04_methodology,sec:noauc}). A feature-fusion head reads the raw features rather than the scores, lies \emph{outside} the theorem, and is audited empirically, with no robust gain~(\cref{sec:06_analysis}).

\paragraph{The reported gains are leakage.}
In our testbed, standard protocols over-credit the more expressive combiners. Leave-one-dataset-out (\lodo) over-credits \emph{routing} because sibling datasets (e.g.\ the Celeb-DF variants~\cite{li2020celeb,li2025celeb}) leak into training, and leave-one-\emph{manipulation}-out (\lomo) over-credits \emph{feature fusion} because the held-out manipulation's source identities and sibling manipulations remain in-distribution~\cite{roessler2019faceforensicspp}. A fold $\times$ eval $2 \times 2$~(\cref{sec:2x2}) isolates the feature-fusion effect as a pure main-effect of \emph{fold} granularity: the $+2.6$pp ``win'' is $12/12$ significant under leaky manipulation folds versus $0/12$ under the honest matched cells~(\cref{tab:supp_2x2}). Each effect forms a \emph{leakage ladder}: re-run under protocols ordered from leakiest to most honest, its point estimate falls monotonically and collapses to noise under leave-one-dataset-\emph{family}-out (\lofo); both hold at every seed, and the feature-fusion artifact also across expert subsets and frame counts~(\cref{sec:generality}).

\paragraph{What actually helps: the tail.}
The corner theorem is silent about the tail, where fusion and expert \emph{diversity} pay off~\cite{dietterich2000ensemble}. As the number of experts grows from 1 to 4, the \allwrong{} rate (videos that every expert misclassifies and that are unrescuable by any combiner) drops from $15.2\%$ to $2.3\%$. Under \lofo, \invb{} beats the best single expert on worst-family $\auc$ ($+4.2$pp), selective risk ($\aurc$, $-2.0$pp)~\cite{geifman2017selective,elyaniv2010foundations}, and low-FPR TPR ($+21.2$pp), and these gains are \emph{protocol-stable}, unlike the $\auc$ ``wins'' that reverse.

\paragraph{Contributions.}
\textbf{(1)} A \emph{corner theorem}: the $\auc$-optimal score-space rule is a label-aware oracle no label-blind router attains in general, leaving the realisable router-versus-fusion comparison open; empirically, every router we test ties or trails global fusion; feature fusion, beyond the theorem's scope, is tested directly~(\cref{sec:04_methodology,sec:06_analysis}).
\textbf{(2)} A \emph{dual-leakage audit}: the apparent routing and feature-fusion gains are evaluation artifacts, formalised by a fold $\times$ eval $2 \times 2$ and two leakage ladders, and shown to hold at every seed~(\cref{sec:05b_main_results,sec:06_analysis}).
\textbf{(3)} A constructive \emph{tail-robustness} result: fusion and diversity reduce the \allwrong{} tail and improve worst-family $\auc$, selective risk, and low-FPR TPR, significantly and stably~(\cref{sec:tail}).
\textbf{(4)} A practical \emph{recipe}, with code and an evaluation snapshot at \coderepo{}. This is an analysis and evaluation-methodology contribution; we do not claim a new state-of-the-art $\auc$~(\cref{sec:08_limitations}).
 \section{Related Work}
\label{sec:03_related_work}

\paragraph{Cross-domain deepfake detection with frozen experts.}
Detectors trained on one manipulation family transfer poorly to the next, so progress is now measured across datasets and manipulations rather than in-domain~\cite{yan2023deepfakebench,yan2024transcending,yermakov2026deepfake,nguyen2025vulnerability}. A dominant response repurposes frozen self-supervised foundation models as detectors, on the premise that general-purpose features generalise across manipulations: CLIP-based forgery detectors~\cite{cui2025forensics,yermakov2025unlocking,radford2021learning}, the spatial transformer DINOv3~\cite{simeoni2025dinov3,oquab2023dinov2}, the video model V-JEPA\,2.1~\cite{assran2025v,murlabadia2026v,bardes2024revisiting}, the face foundation model FSFM~\cite{wang2025fsfm}, and frequency-domain CNNs that read synthesis fingerprints from DCT statistics rather than RGB pixels~\cite{frank2020leveraging}. Our four experts (DINOv3 spatial, V-JEPA\,2.1 temporal, a FREQ-DCT spectral CNN, and FSFM face-SSL) are deliberately drawn from this same line; unlike adapter-style work that fine-tunes a single backbone~\cite{cui2025forensics,kong2026moe}, we keep every expert frozen and study the \emph{combiner} rather than proposing a new representation, pooling the standard benchmarks~\cite{roessler2019faceforensicspp,li2020celeb,li2025celeb,dolhansky2020deepfake,jiang2020deeperforensics} with many further sets into a $22$-dataset suite (\cref{sec:05a_setup}).

\paragraph{Expert fusion, stacking, and per-video routing.}
Combining detectors is mature: parameter-free averaging and stacked generalisation train a second-level learner on first-level scores~\cite{wolpert1992stacked,dietterich2000ensemble}, deep ensembles remain a strong default~\cite{lakshminarayanan2017simple}, and our mean, stack, and inverse-Brier (\invb) combiners are instances over the same in-fold calibrated experts. Routing each input to a specialist is the mixture-of-experts idea~\cite{jacobs1991adaptive,shazeer2017outrageously}, which has reached forgery detection as token- or adapter-level gating inside one backbone~\cite{kong2026moe,zhang2025devfd,yan2025orthogonal}; our gate instead routes at the granularity of the \emph{whole video} across independently trained frozen backbones, producing soft per-video weights, and is thus a per-video instance of mixtures of experts and of learning-to-defer cascades~\cite{mozannar2020consistent,verma2022calibrated,jitkrittum2023when,mao2024principled,mao2025mastering}. Our corner theorem (\cref{thm:corner}) shows the AUC-optimal target these methods chase is a hard, \emph{label-aware} oracle, so the much-discussed gap to a per-video oracle~\cite{souza2017characterization,cruz2017meta} is structurally unrealisable under a ranking metric. This connects to the classical bipartite-ranking and ROC literature~\cite{menon2016bipartite,reid2011information,berta2024classifier}, though the label-aware corner is not the label-blind ROC convex hull.

\paragraph{Calibration is the surviving combiner gain.}
Routing across heterogeneous experts presupposes comparable score scales, which post-hoc calibration supplies: per-expert isotonic regression~\cite{zadrozny2002transforming} and Platt scaling~\cite{platt1999probabilistic} map raw scores to probabilities, and calibrating before combining is known to improve fused decisions~\cite{bella2012effect,guo2017calibration,niculescumizil2005predicting}. In our audit, a single global recalibration captures most of the gate's apparent edge, which is why \invb is the only combiner advantage that survives honest evaluation.

\paragraph{Error \emph{location} versus error \emph{rate}, and the tail.}
That a realisable router does not help AUC over global fusion is, in our setting, an empirical finding; it sits within a long line distinguishing rank-based objectives from error-rate minimisation, where the optimal rule is defined over score \emph{pairs} rather than a single posterior~\cite{cortes2003auc,agarwal2014surrogate,shao2022asymptotically,shen2020auc,mao2023ranking}. What the theorem does \emph{not} cap is the tail, which selective prediction and worst-group analysis quantify beyond average accuracy~\cite{geifman2017selective,elyaniv2010foundations,heng2026know}: this is exactly where fusion and expert diversity make a difference, consistent with evidence that ensembles help most when members are diverse rather than merely strong~\cite{theisen2023when,cruz2015meta}.

\paragraph{Leakage in cross-domain evaluation.}
Leave-one-dataset-out is the de facto cross-domain protocol, but dataset \emph{families} share source identities and pristine reals (the FaceForensics++ manipulations reuse a common pristine pool~\cite{roessler2019faceforensicspp} and Celeb-DF reuses celebrity sources across its versions~\cite{li2020celeb,li2025celeb}), so holding out one sibling while training on the rest leaks low-level statistics into the ``unseen'' manipulation. A second, symmetric hazard is leave-one-\emph{manipulation}-out, where the same dataset's reals and sibling forgeries remain in training, allowing a feature-fusion probe to adapt within the dataset. We quantify both (\cref{sec:06_analysis}); the inflation is a granularity artifact, making leave-one-dataset-\emph{family}-out the honest test of novel-manipulation generalisation.
 \section{A Corner Theorem for Routing}
\label{sec:04_methodology}
We treat cross-domain deepfake detection as a \emph{combiner} problem over a fixed set of frozen experts, and we characterise the \auc{} ceiling of per-video routing. We fix notation, state the combiners, and prove a \emph{corner theorem}, a simple consequence of the ordinal nature of \auc{} whose value is to locate the gap: the \auc{} optimum reachable by routing the expert scores is a hard, label-aware oracle, a ceiling no label-blind rule attains in general. The theorem leaves realisable routers and global combiners unordered, so it is not in itself evidence against routing; that case is empirical (\cref{sec:05b_main_results}).

\paragraph{Setup and combiners.}
Let \K experts produce per-video probabilities $s_1(x),\dots,s_K(x)\in[0,1]$ for a video $x$ with label $y\in\{0,1\}$, calibrated in-fold by isotonic regression~\cite{zadrozny2002transforming} so that no expert sees its evaluation fold (\cref{sec:05a_setup}); write $s(x)=(s_1(x),\dots,s_K(x))$. A \emph{combiner} maps $s(x)$ to a single decision score. A \emph{global} combiner applies a fixed, input-independent rule (the naive mean $\bar s(x)=\tfrac1K\sum_k s_k(x)$, a logistic stack~\cite{wolpert1992stacked}, or inverse-Brier weighting \invb that weights each expert by its in-fold reliability~\cite{lakshminarayanan2017simple,brier1950verification}). In contrast, a \emph{router} applies input-dependent weights $w(x)$ on the probability simplex and outputs the convex mixture $\langle w(x),s(x)\rangle$, the mixture-of-experts gate~\cite{jacobs1991adaptive,shazeer2017outrageously} instantiated per video. A \emph{soft} router uses any $w(x)$ in the simplex, while a \emph{hard} router places all mass on one expert, $w(x)=e_{k(x)}$, recovering dynamic classifier selection~\cite{ko2008dynamic,souza2017characterization}. Feature fusion, a regularised linear probe over the concatenated frozen features, is the remaining expressive baseline, and it sits \emph{outside} the simplex of expert scores.

\paragraph{The corner theorem.}
The router's output is a convex combination of the experts' own scores and so lies in the interval they span. This single geometric fact, combined with the purely ordinal nature of \auc~\cite{cortes2003auc,menon2016bipartite}, bounds routing by a ceiling that only a label-aware oracle reaches.

\begin{theorem}[Score-space routing is bounded by an unrealisable label-aware ceiling]
\label{thm:corner}
Every router of \K experts outputs, for each video, a convex combination of the expert scores, so its score lies in the interval $[\min_k s_k(x),\,\max_k s_k(x)]$. Over all score functions confined to these intervals, \auc is maximised by the \emph{label-aware corner assignment} that gives each positive its maximal expert score and each negative its minimal expert score; this assignment is hard, so softening the weights buys no \auc beyond the best hard assignment. The corner uses the test label to choose $\max$ versus $\min$, so it is \emph{unrealisable}: no label-blind rule (router or global combiner) attains it in general, and it is \emph{not} the experts' ROC convex hull, a label-blind object it strictly dominates.
\end{theorem}

\begin{proof}[Proof sketch]
\auc depends only on the score \emph{ordering}~\cite{cortes2003auc,menon2016bipartite}, and each routed score lies in $[\min_k s_k,\max_k s_k]$, so replacing every positive by $\max_k s_k$ and every negative by $\min_k s_k$ can only increase \auc; this label-aware corner is hard, so no soft router exceeds it. It reads the label, so it is unrealisable; a realisable router and a global combiner are both label-blind and neither class contains the other, so their comparison is empirical. The full argument and the proper-scoring corollary are in supp.\ \cref{supp:proof}.
\end{proof}

In plain terms: whatever its gate reads (scores, features, or the raw video), a router can only move mass \emph{between} the experts' existing scores, and \auc{} cares only about their ordering, so the best conceivable per-video choice is the test-label oracle's, which a real, label-blind router does not reach in general. Feature fusion is \emph{not} covered: a head on the raw features can in principle encode signal absent from any expert's scalar score, so we test it empirically instead (\cref{sec:06_analysis}; \cref{fig:supp_pipelines}), where its apparent advantage proves to be a fold-granularity artifact. We confirm the ceiling numerically (\cref{sec:05b_main_results}): pooled over all videos, the label-aware hard-corner oracle reaches \auc $0.9997$ at $\K{=}4$, and soft routers match hard routers to within cluster-bootstrap noise. Under a strictly proper score such as Brier~\cite{gneiting2007strictly,brier1950verification}, a soft gate can beat the mean, but a single global recalibration of the mean recovers most of that gap (pooled Brier $0.111{\to}0.092$ versus the gate's $0.090$), so the proper-scoring advantage is a calibration effect rather than a routing one (supp.\ \cref{supp:proof}). This motivates our recipe: calibrate, then combine \emph{globally}.

\paragraph{Hard routing loses to global fusion, and not for want of a stronger gate.}
The usual defence is that a stronger gate would eventually win, so we measure the ceiling it assumes: the \emph{winner-predictability} $q$, the out-of-family cross-validated accuracy of the best predictor of the per-video oracle winner fit on the router's own features. We find $q=0.84$ against a $0.39$ four-way majority baseline, so the oracle winner is in fact highly predictable; the gate is not the bottleneck. Yet hard-routing that \emph{predicted} winner scores family-macro \auc $0.940$, significantly \emph{below} the calibrated global combiner \invb (paired $\Delta{=}-0.79$pp, $95\%$ CI $[-1.1,-0.4]$; \lofo, seed~$42$). Hard-routing the predicted winner therefore loses to global fusion even though the winner is predictable: committing to one expert discards the cross-expert agreement a global combiner keeps, so the shortfall is not a weak-gate artifact.
 \section{Experimental Setup}
\label{sec:05a_setup}

\paragraph{Experts.}
We combine four architecturally diverse frozen experts, each a foundation model with a lightweight, trained head, each emitting one calibrated probability per video. \textbf{DINOv3} is a spatial expert (ViT-L self-supervised features~\cite{simeoni2025dinov3}), adapted with an orthogonal-pair gradient-target LoRA objective~\cite{collins2026paired,hu2022lora} and scored over $32$ frames spread across the clip duration. \textbf{V-JEPA\,2.1} is a temporal expert (ViT-L video features~\cite{murlabadia2026v,assran2025v}) fine-tuned with supervised contrastive training and extended LoRA~\cite{khosla2020supervised,hu2022lora} and scored over contiguous frames. \textbf{FREQ-DCT} is a spectral DCT-domain CNN in the spirit of frequency-based forensics~\cite{frank2020leveraging}; it is weak alone (${\sim}0.68$ standalone \auc) but spectrally orthogonal to the other three, exactly the diversity the tail rewards. \textbf{FSFM} is a face self-supervised ViT-L, whose released face-security detector~\cite{wang2025fsfm} we run unchanged on the same manifests and with the same aggregation. All four are trained only on the FaceForensics++ c23 manipulations~\cite{roessler2019faceforensicspp} and evaluated zero-shot on every other dataset; DINOv3, V-JEPA, and FREQ-DCT are trained per seed under a matched recipe, while FSFM uses its native checkpoint (scored once and reused). The pool is fixed \emph{a priori}, one expert per artifact cue of \cref{sec:02_introduction}, with no subset search on test performance; results span the nested subsets $\K{=}2..4$ plus a fifth expert (supp.\ \cref{supp:tailk5}), and two members (FREQ-DCT, FSFM) are deepfake-specific detectors rather than generic backbones. The experts are held fixed throughout, each backbone scored at its native window rather than via an adaptive frame sampler~\cite{wu2019adaframe,zhi2021mgsampler,ghodrati2021frameexit,tang2025adaptive}, so every claim concerns the combiner on a \emph{fixed} score matrix and does not hinge on reproducing a given expert's training~\cite{collins2026paired}.

\paragraph{Datasets.}
Evaluation spans $22$ video deepfake test sets covering face-swap, face-reenactment, talking-head, and lip-sync manipulations across graphics- and learning-based generation, drawn from FaceForensics++~\cite{roessler2019faceforensicspp}, the DeepFake Detection Challenge dataset and its preview split~\cite{dolhansky2020deepfake}, the Google/Jigsaw DeepFake Detection set~\cite{dufour2019contributing}, the face-swap attribution set DFDM~\cite{jia2022model}, FakeAVCeleb~\cite{khalid2021fakeavceleb}, KoDF~\cite{kwon2021kodf}, the Celeb-DF series including its third release Celeb-DF++ (v3 in our tables)~\cite{li2020celeb,li2025celeb}, the DeepSpeak series~\cite{barrington2026deepspeak}, PolyGlotFake~\cite{hou2024polyglotfake}, IDForge~\cite{xu2024identity}, DeeperForensics-1.0~\cite{jiang2020deeperforensics}, UADFV~\cite{yang2019exposing}, and Face Forensics in the Wild~\cite{zhou2021face}. We pool datasets that share source identities and pristine reals into \emph{families}, the held-out unit of our honest protocol. Membership is fixed as follows: FF++ (its six c23 subsets, pristine and manipulated), CDF (Celeb-DF v1/v2/v3 and DFDM, whose reals are Celeb-DF reals), DFDC (DFDC and its preview DFDCP), and DS (DeepSpeak v1/v2) are the four multi-member families, and every other dataset is its own family, giving twelve held-out units (\cref{tab:perfamily}).

DFDM joined CDF after a post-hoc audit of our own grouping caught it sharing pristine reals verbatim with the Celeb-DF series (48 of Celeb-DF-v2's 178 reals), the very sibling leakage the protocol exists to remove; every number in the paper uses this corrected grouping. The one external borderline case, DeeperForensics-1.0, reuses the FaceForensics++ reals~\cite{jiang2020deeperforensics}; we keep it a separate, leakier unit, and the honest verdict, on ranking and on every tail metric, is unchanged under coarser or finer definitions (supp.\ \cref{supp:familyrobust,supp:tailfamilyrobust}). Every expert is trained on FF++ c23, yet FF++ is also an evaluation family; a check without it (eleven families), whose readout rule we fixed and committed before running it, leaves the conclusions unchanged: under \lofo, \invb$-$mean stays $+0.27$pp (CI excludes $0$), gate$-$mean is not significantly positive, and every tail claim keeps its sign and significance, while absolute AUC and TPR levels fall slightly (supp.\ \cref{supp:ffppexcluded}).

\paragraph{Combiners.}
Using the same calibrated score matrix and the same folds, we compare five combiners. Hence, any difference reflects \emph{how} experts are weighted rather than differences in calibration. \emph{Best-single} is the expert with the best out-of-fold macro \auc{}, picked per seed (our reference baseline, almost always DINOv3); \emph{mean} fusion is the naive parameter-free average; the \emph{stack} is a global logistic stacker~\cite{wolpert1992stacked}; \invb is a global ensemble that weights experts by inverse in-fold Brier~\cite{lakshminarayanan2017simple,brier1950verification}; and the \emph{gate} is a per-video router: a gradient-boosted classifier that predicts the per-video oracle-winner expert (the expert with the lowest per-video cross-entropy) from the calibrated scores, their confidences $|s_k{-}0.5|$, and all pairwise disagreements $|s_i{-}s_j|$, $14$ inputs at $\K{=}4$, and whose soft class-probabilities act as routing weights~\cite{shazeer2017outrageously,mao2025mastering}. \emph{Feature fusion} is the separate, strongly regularised linear probe of \cref{sec:04_methodology}. Every combiner is fit strictly out-of-fold with in-fold isotonic calibration; model classes, hyperparameters, and fold handling are in supp.\ \cref{supp:impl}.

\paragraph{Per-video scores.}
We mean-pool each expert's per-frame predictions to one score per video and join experts on \texttt{(dataset, method, video\_id, label)} (pipeline: supp.\ \cref{fig:supp_flow}).\label{par:join} The manipulation method is part of the key because a source \texttt{video\_id} is shared across a dataset's manipulations, so each manipulation stays distinct, and the label keeps paired-identity datasets two-class.

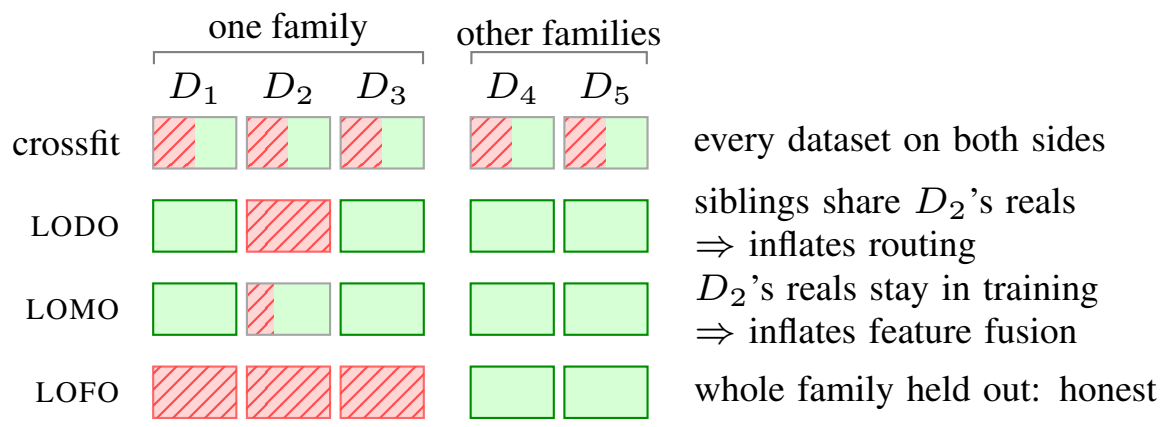
\begin{figure}[t]
\centering
\begin{tikzpicture}[x=1mm,y=1mm,font=\scriptsize]
\def\w{2.75}\def\h{1.7}
\newcommand*{\pcTr}[2]{\filldraw[fill=green!16,draw=green!55!black] (#1-\w,#2-\h) rectangle (#1+\w,#2+\h);}
\newcommand*{\pcTe}[2]{\filldraw[fill=red!16,draw=red!60] (#1-\w,#2-\h) rectangle (#1+\w,#2+\h);
  \fill[pattern=north east lines, pattern color=red!70] (#1-\w,#2-\h) rectangle (#1+\w,#2+\h);}
\newcommand*{\pcSplit}[3]{\fill[red!16] (#1-\w,#2-\h) rectangle (#1-\w+#3,#2+\h);
  \fill[pattern=north east lines, pattern color=red!70] (#1-\w,#2-\h) rectangle (#1-\w+#3,#2+\h);
  \fill[green!16] (#1-\w+#3,#2-\h) rectangle (#1+\w,#2+\h);
  \draw[gray!70] (#1-\w,#2-\h) rectangle (#1+\w,#2+\h);}
\foreach \x/\n in {14.5/1,20.7/2,26.9/3,35.5/4,41.7/5} \node[font=\scriptsize] at (\x,3.6) {$D_\n$};
\draw[gray] (11.8,5.2) -- (11.8,6) -- (29.6,6) -- (29.6,5.2);
\node[font=\scriptsize, above, inner sep=1pt] at (20.7,6) {one family};
\draw[gray] (32.8,5.2) -- (32.8,6) -- (44.4,6) -- (44.4,5.2);
\node[font=\scriptsize, above, inner sep=1pt] at (38.6,6) {other families};
\node[anchor=east] at (10.8,0)   {crossfit};
\node[anchor=east] at (10.8,-5.5)  {\lodo};
\node[anchor=east] at (10.8,-11) {\lomo};
\node[anchor=east] at (10.8,-16.5) {\lofo};
\foreach \x in {14.5,20.7,26.9,35.5,41.7} {\pcSplit{\x}{0}{\w}}
\pcTr{14.5}{-5.5}\pcTe{20.7}{-5.5}\pcTr{26.9}{-5.5}\pcTr{35.5}{-5.5}\pcTr{41.7}{-5.5}
\pcTr{14.5}{-11}\pcSplit{20.7}{-11}{1.85}\pcTr{26.9}{-11}\pcTr{35.5}{-11}\pcTr{41.7}{-11}
\pcTe{14.5}{-16.5}\pcTe{20.7}{-16.5}\pcTe{26.9}{-16.5}\pcTr{35.5}{-16.5}\pcTr{41.7}{-16.5}
\begin{scope}[every node/.style={anchor=west, align=left, text width=35mm, font=\scriptsize}]
\node at (46.3,0)   {every dataset on both sides};
\node at (46.3,-5.5)  {siblings share $D_2$'s reals\\$\Rightarrow$ inflates routing};
\node at (46.3,-11) {$D_2$'s reals stay in training\\$\Rightarrow$ inflates feature fusion};
\node at (46.3,-16.5) {whole family held out: honest};
\end{scope}
\end{tikzpicture}
\caption{\textbf{Protocols and their leakage.} Each row splits five datasets into combiner training (green) and test (red, hatched); $D_1$--$D_3$ share reals, and under \lomo{} the test set is one manipulation of $D_2$. Crossfit, \lodo{} and \lofo{} form the leakage ladder of \cref{tab:ladder}. Full version: supp.\ \cref{fig:supp_leakage}.}
\label{fig:protocols}
\end{figure}

\paragraph{Protocols.}
We evaluate under cross-validation protocols that increase honesty along two axes, because the two expressive combiners leak along different axes (\cref{sec:06_analysis}). On the \emph{dataset} axis: \emph{crossfit} (random stratified folds, with same-dataset videos in both train and test), \lodo (leave-one-dataset-out, under which sibling datasets such as Celeb-DF-v1/v2/v3 still leak), and \lofo (leave-one-dataset-\emph{family}-out, which holds an entire family out at once and is our honest cross-domain protocol). On the \emph{manipulation} axis: \lomo (leave-one-manipulation-out), under which the held-out manipulation's source reals and sibling manipulations remain in-distribution, enabling within-dataset adaptation (\cref{fig:protocols}). Unless stated otherwise, the headline protocol is \lofo, and every reported number is a mean over three seeds ($42,137,1729$): single-seed evaluation is itself a pitfall we document (\cref{sec:generality}).

\paragraph{Metric and significance.}
Our primary metric is family-macro \auc (the mean of per-family video-\auc, skipping single-class families so that small and large families contribute equally), and we also report the video-weighted (micro) \auc. \auc and the low-FPR points are rank-based and hence invariant under strictly monotone recalibration, thereby isolating ranking from calibration~\cite{menon2016bipartite}; for calibration quality, we use the strictly proper Brier score~\cite{gneiting2007strictly,brier1950verification}. For the tail, where \cref{thm:corner} imposes no cap, we report the \allwrong rate (the fraction of videos every expert misclassifies at $0.5$ on out-of-fold calibrated scores), worst-family \auc, the area under the risk--coverage curve (\aurc, selective risk~\cite{geifman2017selective}), and TPR at $1\%$ FPR. All deltas carry $95\%$ paired cluster-bootstrap confidence intervals that resample \emph{families}, and we call a result significant only when the interval excludes zero.
 \section{Main Results}
\label{sec:05b_main_results}

\begin{table}[t]
\centering
\caption{\textbf{Combiner comparison (\lofo, the honest cross-domain protocol).}
The four experts (top) span a wide \auc{} range, yet every fused score-space combiner (bottom) lands in a narrow $0.946$--$0.949$ macro-\auc{} band, far below the label-aware ceiling (\cref{thm:corner}); the \emph{feature-fusion} probe sits outside that score space and does not yield a robust, protocol-stable improvement over \invb ($+0.28$pp, significant at no seed; \cref{sec:06_analysis}).
Brier is the deployed recipe (calibrate in-fold, combine, recalibrate). mean$\pm$std over three seeds (FSFM single frozen seed). Dashes ($-$) mark metrics not computed: experts and the probe are scored on \auc{} only.
Best-single is evaluated on the same calibrated out-of-fold scores as the combiners, hence its small offset from the raw-score expert rows. Bold: best score-space combiner per column.}
\label{tab:main}
\setlength{\tabcolsep}{4pt}
\resizebox{\linewidth}{!}{\begin{tabular}{lccccc}
\toprule
 & macro \auc & micro \auc & pAUC@5\% & TPR@1\%FPR & Brier$\downarrow$ \\
\midrule
\textbf{DINOv3}      & $0.931{\pm}0.003$ & $0.919{\pm}0.001$ & -- & -- & -- \\
\textbf{V-JEPA\,2.1} & $0.917{\pm}0.001$ & $0.908{\pm}0.002$ & -- & -- & -- \\
\textbf{FREQ-DCT}    & $0.680{\pm}0.011$ & $0.673{\pm}0.024$ & -- & -- & -- \\
\textbf{FSFM}        & $0.895$           & $0.887$           & -- & -- & -- \\
\midrule
best-single (DINOv3) & $0.930{\pm}0.003$ & $0.908{\pm}0.002$ & $0.813{\pm}0.007$ & $0.567{\pm}0.022$ & $0.111{\pm}0.001$ \\
\emph{mean}          & $0.947{\pm}0.001$ & $0.933{\pm}0.001$ & $0.840{\pm}0.003$ & $0.621{\pm}0.012$ & $0.099{\pm}0.001$ \\
\emph{stack}         & $\mathbf{0.949{\pm}0.001}$ & $\mathbf{0.935{\pm}0.001}$ & $\mathbf{0.851{\pm}0.002}$ & $\mathbf{0.656{\pm}0.006}$ & $\mathbf{0.096{\pm}0.001}$ \\
\invb{} (recipe)     & $\mathbf{0.949{\pm}0.001}$ & $0.934{\pm}0.001$ & $0.849{\pm}0.001$ & $0.648{\pm}0.004$ & $0.097{\pm}0.001$ \\
\emph{gate} (per-video) & $0.946{\pm}0.001$ & $0.928{\pm}0.001$ & $0.846{\pm}0.005$ & $0.631{\pm}0.013$ & $0.100{\pm}0.001$ \\
\emph{feature-fusion} & $0.952{\pm}0.003$ & $0.946{\pm}0.003$ & -- & -- & -- \\
\bottomrule
\end{tabular}}
\end{table}
 \begin{table*}[t]
\centering
\caption{\textbf{Per-family cross-dataset video \auc{} (\lofo).}
Each cell pools every video in a manipulation family and scores one expert's
out-of-fold predictions; the recipe row (\invb) is the inverse-Brier combination of the
four calibrated experts.
Family columns: CDF (Celeb-DF v1/v2, Celeb-DF++ (v3), and DFDM, which shares its pristine reals),
DFDC (DFDC/DFDCP), DS (DeepSpeak v1/v2), FFpp (the six FaceForensics++ subsets), and
eight singleton datasets; Macro is the equal-weight family mean.
Mean over three seeds (FSFM is a single frozen seed).
Best expert per family in \textbf{bold}.
The experts are complementary rather than redundant: no single expert dominates
(V-JEPA leads CDF, DS, FFpp, and FFIW; DINOv3 leads six families and the Macro; FSFM
leads DFDC and UADFV), and the calibrated \invb{} combination exceeds or matches (within
$0.5$pp) the best expert on eleven of twelve families and on Macro, the exception being
DeeperForensics, where the spatial expert alone excels.
FREQ-DCT leads no family; its contribution is the tail, where it shrinks the \allwrong{}
blind spot most (\cref{sec:tail}).}
\label{tab:perfamily}
\setlength{\tabcolsep}{4pt}
\resizebox{\textwidth}{!}{\begin{tabular}{lccccccccccccc}
\toprule
expert & CDF & DFDC & DS & FFpp & DFD & DeFo & FAVC & FFIW & IDFv1 & KoDF & PGF & UADFV & Macro \\
\midrule
\textbf{DINOv3}      & $0.855$ & $0.867$ & $0.856$ & $0.982$ & $\mathbf{0.973}$ & $\mathbf{0.990}$ & $\mathbf{0.980}$ & $0.885$ & $\mathbf{0.967}$ & $\mathbf{0.903}$ & $\mathbf{0.932}$ & $0.977$ & $\mathbf{0.931}$ \\
\textbf{V-JEPA\,2.1} & $\mathbf{0.909}$ & $0.806$ & $\mathbf{0.900}$ & $\mathbf{0.985}$ & $0.964$ & $0.868$ & $0.978$ & $\mathbf{0.890}$ & $0.959$ & $0.879$ & $0.910$ & $0.954$ & $0.917$ \\
\textbf{FREQ-DCT}    & $0.719$ & $0.600$ & $0.588$ & $0.682$ & $0.586$ & $0.543$ & $0.720$ & $0.625$ & $0.753$ & $0.706$ & $0.705$ & $0.938$ & $0.680$ \\
\textbf{FSFM}        & $0.882$ & $\mathbf{0.883}$ & $0.816$ & $0.962$ & $0.913$ & $0.962$ & $0.912$ & $0.839$ & $0.858$ & $0.855$ & $0.867$ & $\mathbf{0.988}$ & $0.895$ \\
\midrule
\invb{} (recipe)     & $0.930$ & $0.896$ & $0.903$ & $0.987$ & $0.969$ & $0.971$ & $0.990$ & $0.923$ & $0.976$ & $0.908$ & $0.951$ & $0.987$ & $0.949$ \\
\bottomrule
\end{tabular}}
\end{table*}
 
\cref{tab:main} lays out the comparison: four heterogeneous experts and six ways to combine them, evaluated across all 22 datasets under the honest \lofo protocol. The experts span a wide range (FREQ-DCT $0.680$ to DINOv3 $0.931$ family-macro \auc, a $25$pp spread) and are complementary, three of the four winning families outright while the fourth earns its seat in the tail (\cref{tab:perfamily}); yet the fused score-space combiners collapse into a narrow $0.946$--$0.949$ macro-\auc{} band, and even the feature-fusion probe, though outside the theorem's score space, yields no robust improvement over them under honest folds. The expert \emph{set} carries almost all of the ranking signal and the combiner almost none; \invb and stack lead on Brier.

\subsection{No robust expressive-combiner AUC gain under LOFO}
\label{sec:noauc}
Under \lofo, the expressive combiners do not beat global fusion on family-macro AUC. This is an empirical finding: \cref{thm:corner} only places the routing optimum at a label-aware oracle. The per-video gate ties mean fusion (gate$-$mean $=-0.07$pp, n.s.; gate ahead at $0/3$ seeds); a paired cluster-bootstrap two-one-sided-test (TOST) over families upgrades this from a mere non-rejection to a \emph{powered} equivalence, certifying $|\Delta|<0.5$pp ($p{=}0.002$) at a minimum detectable effect of $0.39$pp (supp.\ \cref{supp:equivalence}). A \emph{content-conditioned} router, its gate reading a frozen video-language embedding instead of the scores, does no better: soft variants tie \invb ($+0.03$/$+0.12$pp, n.s.\ in every seed) and hard variants are significantly worse ($-2.2$pp; supp.\ \cref{supp:contentrouter}). Feature fusion is by contrast \emph{not} a robust, protocol-stable win over \invb under honest folds ($+0.28$pp $3$-seed, significant at no seed); the held-out family count is too small to certify \emph{equivalence} there (per-seed MDE ${\approx}1.9$--$2.5$pp), so that verdict rests instead on the fold$\times$eval decomposition (\cref{sec:06_analysis}). The surviving combiner effect is calibration: \invb beats mean fusion by $+0.26$--$0.29$pp, significant in \emph{all} protocols (\cref{tab:ladder}). The gain comes from global recalibration~\cite{bella2012effect} and is small because calibrated mean fusion is already near-optimal for AUC~\cite{niculescumizil2005predicting}. The apparent superiority of routing is a leakage artifact: as the dataset-axis protocol is made honest, the gate$-$mean point estimate falls monotonically from $+0.36$pp (crossfit, significant) to $+0.16$pp (\lodo, not significant, though the gate leads at all three seeds) to $-0.07$pp (\lofo; \cref{tab:ladder}).

\begin{table}[t]
\centering\small
\setlength{\tabcolsep}{4pt}
\begin{tabular}{lccc}
\toprule
macro AUC, $\Delta$ vs.\ mean & crossfit & \lodo & \lofo \\
\midrule
gate $-$ mean (routing)     & $+0.36^{*}$ & $+0.16$ & $-0.07$ \\
\invb $-$ mean (calibration) & $+0.26^{*}$ & $+0.29^{*}$ & $+0.27^{*}$ \\
\bottomrule
\end{tabular}
\caption{\textbf{Routing leakage ladder.} The gate's point estimate shrinks and reverses as the dataset-axis protocol becomes honest (\lodo: n.s., gate ahead at $3/3$ seeds), while \invb's calibration gain is protocol-stable; the honest (\lofo) gate$-$mean tie is a \emph{powered} equivalence (supp.\ \cref{supp:equivalence}). $^{*}$: $95\%$ cluster-bootstrap CI excludes $0$ ($3/3$ seeds).}
\label{tab:ladder}
\end{table}

\subsection{What fusion \textbf{\emph{does}} buy: the tail}
\label{sec:tail}
The corner theorem does not constrain the \emph{tail}, and there heterogeneous experts and global fusion earn gains that hold under every protocol (\cref{tab:tail}).

\paragraph{Diversity law.}
The \allwrong rate, beyond any combiner's reach, falls monotonically as experts are added: $15.2\%\to7.1\%\to4.6\%\to2.3\%$ for $\K{=}1{\to}4$. The drop from the best pair to the full set ($K2{-}K4$) is $+4.7$pp and significant; a complementary expert shrinks the unrecoverable tail roughly six-fold, the practical form of the classical finding that ensemble benefit flows from error diversity~\cite{theisen2023when,dietterich2000ensemble,lakshminarayanan2017simple}, a property of the expert \emph{set} alone.

\paragraph{Fusion robustness.}
Calibrated global fusion (\invb) also beats the single best expert wherever robustness is measured: worst-family AUC ($+4.2$pp), selective risk (\aurc, $-2.0$pp; lower is better)~\cite{geifman2017selective,elyaniv2010foundations}, and TPR at $1\%$ FPR ($+21.2$pp), all significant under \lofo. Unlike the AUC ``wins'' of \cref{tab:ladder}, these tail gains are \emph{stable across leaky and honest protocols}: they neither inflate under crossfit/\lodo{} nor vanish under \lofo{} (\cref{tab:tail}), the signature of a real effect rather than leakage. For practitioners adding an expert, the tail is therefore a better success metric than macro AUC~\cite{cortes2003auc}.

\begin{table}[t]
\centering\small
\setlength{\tabcolsep}{4pt}
\begin{tabular}{lccc}
\toprule
 & \lofo & crossfit & \lodo \\
 & (honest) & (leaky) & (leaky) \\
\midrule
\allwrong $K2{-}K4$ ($\downarrow$tail) & $+4.7^{*}$ & $+4.5^{*}$ & $+4.9^{*}$ \\
worst-family AUC (\invb$-$best)        & $+4.2^{*}$ & $+4.2^{*}$ & $+6.7^{*}$ \\
\aurc, selective risk ($-$, better)     & $-2.0^{*}$ & $-1.7^{*}$ & $-2.0^{*}$ \\
TPR@$1\%$ FPR (\invb$-$best)            & $+21.2^{*}$ & $+15.8^{*}$ & $+21.4^{*}$ \\
\bottomrule
\end{tabular}
\caption{\textbf{The tail-robustness positive (pp).} Fusion and diversity help the tail \emph{significantly and protocol-stably}, the opposite of the leaky AUC ``wins'' in \cref{tab:ladder}. $^{*}$: CI excludes $0$. \allwrong curve $\K{=}1{\to}4$: $15.2{\to}7.1{\to}4.6{\to}2.3\%$.}
\label{tab:tail}
\end{table}

\paragraph{A prediction for new experts, borne out.}
A matched CLIP/X-CLIP~\cite{radford2021learning,ni2022expanding} fifth expert confirms the diversity-law prediction at all three seeds (tail $2.3\%{\to}1.8\%$, $K2{\to}K5$ $+5.3^{*}$pp) while the gate turns significantly \emph{worse} than mean fusion ($-0.40$pp, CI $[-0.77,-0.03]$; supp.\ \cref{supp:tailk5}).
 \section{The Evaluation Is Leaky}
\label{sec:06_analysis}
The apparent gains of both expressive combiners are artifacts of training-fold granularity. Both artifacts appear at every seed, and the feature-fusion one persists across expert subsets and frame counts (\cref{fig:supp_bars}).

\subsection{Feature fusion: a fold$\times$eval $2{\times}2$}
\label{sec:2x2} 
\begin{figure}[t]
\centering
\includegraphics[width=\linewidth]{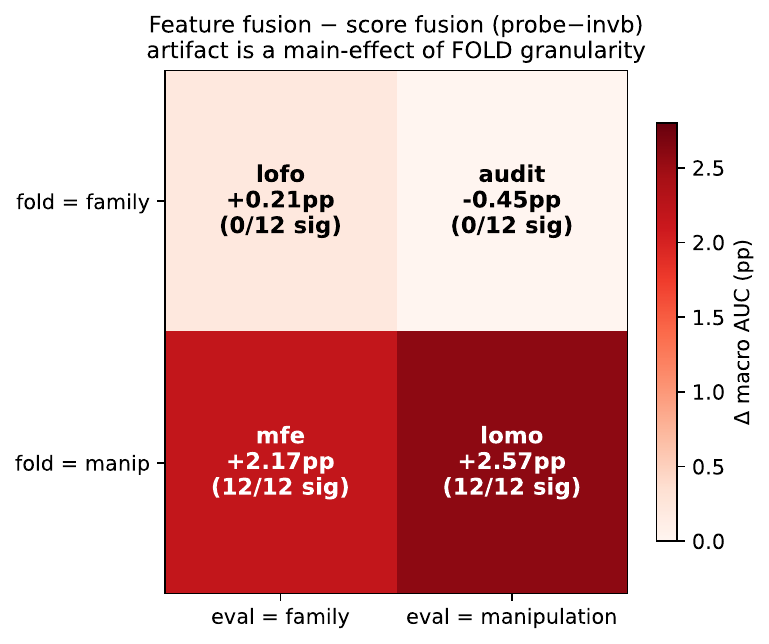}
\caption{\textbf{The artifact is a main effect of FOLD granularity.} Feature fusion $-$ score fusion (probe$-$\invb macro AUC, pp; grand mean over subsets and seeds, \#sig$/12$). Rows vary the training-fold granularity; columns only the evaluation granularity. The effect follows the rows and barely changes across the columns, so the gain comes from leaky training folds rather than the metric. The honest-fold row is $\sim 0$ ($+0.21$/$-0.45$, $0/12$); the manipulation-fold row is $+2.2$--$2.6$pp and $12/12$ significant at \emph{both} eval granularities.}
\label{fig:2x2}
\end{figure}

The feature-fusion probe beats \invb by $+2.57$pp under leave-one-manipulation-out (\lomo), which looks like a decisive joint-fusion capability beyond score fusion~\cite{cui2025forensics,das2023unmasking}. To test it, we decouple the two granularities \lomo conflates, the \emph{fold} (training leakage) and the \emph{eval} (macro grouping), into a $2{\times}2$ of probe$-$\invb (grand mean over subsets and seeds, \#sig$/12$; \cref{fig:2x2}). Switching the \emph{fold} from family to manipulation flips probe$-$\invb from $+0.21$ ($0/12$) to $+2.17$ ($12/12$) at the family eval and from $-0.45$ ($0/12$) to $+2.57$ ($12/12$) at the manipulation eval; switching the \emph{eval} granularity at a fixed fold barely moves it. The inflation is therefore a clean main-effect of fold granularity (within-dataset \emph{adaptation}: the fold holds out the manipulation but not its dataset~\cite{yan2024transcending}), and not a real manipulation-weighting capability. Under honest folds, feature fusion shows no robust improvement over score fusion, and this holds under a nested-CV \emph{tuned} probe (honest \lofo{} $+0.35$pp, $0/3$ seeds).

\subsection{A symmetric leakage ladder}
\label{sec:ladder}
On the dataset axis, holding eval at the honest (family) level and walking the \emph{fold} from leaky to honest, probe$-$\invb forms a monotone ladder (crossfit $+3.71$ $[3/3]>$ \lodo $+1.19$ $[3/3]>$ \lofo $+0.28$ $[0/3]$), the same story \cref{tab:ladder} told for routing. Thus \emph{both} expressive mechanisms leak on \emph{both} axes: routing via sibling-dataset leakage on the dataset axis (\cref{sec:noauc}), feature fusion via within-dataset adaptation on the manipulation axis, and both collapse to noise under \lofo. This dual-leakage audit (per-video join, \cref{par:join}) shows how standard cross-dataset~\cite{yan2023deepfakebench} and cross-manipulation~\cite{roessler2019faceforensicspp} protocols can over-credit expressive combiners.

\subsection{Generality and the single-seed pitfall}
\label{sec:generality}
The $2{\times}2$ pattern is not specific to the four-expert set: across all four expert subsets ($\K{=}2\!\to\!4$, spanning two ViTs~\cite{dosovitskiy2021image}, ViT$+$CNN, and ViT$+$face) and three seeds it holds with $12/12$ significant manipulation-fold cells (\lomo) and $0/12$ honest-fold cells at \emph{either} eval granularity (per-subset breakdown in \cref{fig:supp_bars}), and it is frame-count robust: a $64$-frame re-score reproduces the pattern, honest-fold cells null, manipulation-fold cell significant. A third pitfall emerges in passing: the honest-fold edge is carried almost entirely by seed $42$ ($+0.70$ versus $+0.03$/$+0.11$pp), and under the pre-audit grouping that single seed reached significance ($+0.96$pp) where the three-seed verdict does not: multi-seed reporting is not optional~\cite{lakshminarayanan2017simple}.
 \section{Discussion}
\label{sec:07_discussion}
\paragraph{A recipe.}
Our recipe for combining frozen experts: calibrate each expert in-fold, combine globally with inverse-Brier weighting (\invb), evaluate leave-one-dataset-\emph{family}-out, and report tail metrics (\allwrong, worst-family AUC, selective risk \aurc) alongside AUC. This needs no per-video gate and no concatenated-feature fuser (\cref{sec:noauc,sec:06_analysis}); \invb is the only combiner that beats mean fusion significantly under every protocol.

\paragraph{Where progress comes from.}
Since score-space routing does not beat global fusion on \auc{} under \lofo{} in our tests (\cref{sec:noauc}), the only levers left are the \emph{expert set} and the \emph{tail}, which \cref{thm:corner} leaves uncapped. The diversity law (\cref{sec:tail}) makes it concrete: an architecturally orthogonal expert helps exactly when it rescues videos the current set gets unanimously wrong, which the \allwrong overlap measures a priori and macro AUC misses, since ensemble gains depend more on diversity than on individual strength~\cite{theisen2023when,cruz2015meta}; our weak FREQ-DCT expert moves the average least and the blind spot most. It reframes ``which combiner?'' as ``which expert to add, and does it shrink the tail?''.

\paragraph{When routing might genuinely help.}
The evidence against routing is limited to frozen experts combined in score space under rank- or threshold-based objectives. Routing can pay off when experts have disjoint or out-of-score-space competence (deferral to an unseen modality, abstention) or when the objective is not rank-based (selective prediction with a hard reject, asymmetric per-domain costs)~\cite{mozannar2020consistent,verma2022calibrated,mao2025mastering,shen2020auc}; our frozen experts share one calibrated score space and overlap heavily, so none of these escape hatches applies, and for AUC and low-FPR operating points we recommend fusion over routing.

\paragraph{Implications for the field.}
Fusion and routing ``wins'' under crossfit, \lodo, or \lomo{} rest on protocols that inflate expressive combiners in our testbed; they are not useless, but their margins should be re-measured \lofo, with recalibrated baselines and low-FPR points, before crediting combiner \emph{capability} rather than fold granularity.
 \section{Limitations}
\label{sec:08_limitations}
We contribute analysis and evaluation methodology and claim no state-of-the-art result: we fix the experts to study the combiner, and our conclusions are scoped to homogeneously scored, overlapping frozen experts. Training every trainable expert on the same FaceForensics++ c23 corpus is a further deliberate control, isolating the combiner on a fixed competence profile; a pool trained on disjoint domains could widen score diversity and give routing more headroom (\cref{sec:07_discussion}). The corner theorem (\cref{thm:corner}) is an \emph{AUC} statement; under Brier, a soft router can differ from a global combiner~\cite{gneiting2007strictly,nguyen2026convergence}, but the proper-scoring corollary (supp.\ \cref{supp:proof}) attributes that gap to calibration. Our negative routing result covers both a score-derived gate ($q{=}0.84$) and a \emph{content-conditioned} gate reading a frozen video-language embedding (supp.\ \cref{supp:contentrouter}); both stay within the theorem's convex hull, as any router must~\cite{mao2025mastering,charusaie2022sample}, and neither beats \invb. We have not tested end-to-end joint fine-tuning, which alters the experts and can leave score space entirely. Honest cross-domain evaluation leaves few held-out units: \lofo removes intra-family leakage but yields only $12$ grouped targets, so we treat it as the honest, deliberately conservative estimate and trust only effects that survive three seeds, family clustering, and alternative family definitions (supp.\ \cref{supp:familyrobust}); no honest-fold feature-fusion cell survives that bar (\cref{sec:generality}). In-fold isotonic calibration and the global \invb weights assume the held-out family resembles the calibration pool in score scale~\cite{zadrozny2002transforming,perezlebel2025decision}; under severe shift, these maps can drift, so we report recalibration-invariant rank metrics alongside proper scores. We cannot exhaustively audit the pretraining corpora of our foundation models (DINOv3, V-JEPA, FSFM, X-CLIP) against the $22$ test sets, a caveat shared by all such detectors. The $22$-dataset suite skews toward GAN/graphics and talking-head manipulations; currency against diffusion and audio-driven fakes~\cite{cai2025av,barrington2026deepspeak} and robustness to social-media re-encoding remain future work; the structural AUC ceiling holds regardless, though per-expert and tail numbers may shift.
 \section{Conclusion}
\label{sec:09_conclusion}
For honest cross-domain $\auc$, the combiner barely matters: the corner theorem places the optimum of score-space routing at a label-aware oracle, no realisable router we test beats globally calibrated \invb{} under \lofo, and feature fusion, outside the theorem, adds no robust, protocol-stable gain. Fusion pays in the uncapped \emph{tail} (\allwrong{} $15.2\%{\to}2.3\%$): calibrate, combine globally, evaluate \lofo, and measure tail success.
 
\section*{Acknowledgements}
This research was supported by a Deakin University Postgraduate Research (DUPR) Scholarship.

{\small
\bibliographystyle{ieeenat_fullname}

}

\twocolumn[{\centering\Large\bfseries Supplementary Material\par}\vspace{1.5em}]
\setcounter{figure}{0}
\setcounter{table}{0}
\appendix

\renewcommand{\thefigure}{S\arabic{figure}}
\renewcommand{\thetable}{S\arabic{table}}

\section{Schematic Overview}
\label{supp:schematics}
\begin{figure*}[t]
\centering
\begin{tikzpicture}[
  font=\footnotesize,
  >={Stealth[length=1.8mm]},
  base/.style={draw, rounded corners=2pt, align=center, inner sep=4pt, minimum height=8mm},
  data/.style={base, fill=gray!10},
  proc/.style={base, fill=blue!7},
  ff/.style={base, fill=red!15, draw=red!55, thick},
  route/.style={base, fill=orange!22, draw=orange!75, thick},
  glob/.style={base, fill=green!18, draw=green!55!black, thick},
  outp/.style={draw, circle, fill=gray!5, inner sep=1pt, minimum size=7mm},
  ed/.style={draw, ->, gray!65},
]
\node[data, text width=15mm] (vid) {Raw video $x$};
\node[proc, right=8mm of vid, text width=26mm] (exp)
  {Four frozen experts\\[1pt]{\scriptsize DINOv3 (spatial) $\cdot$ V-JEPA (temporal)}\\{\scriptsize FREQ-DCT (spectral) $\cdot$ FSFM (face)}};
\draw[ed] (vid) -- (exp);

\node[base, fill=red!6, above right=7mm and 14mm of exp, text width=21mm] (concat) {Concatenate\\raw features};
\node[ff, right=7mm of concat, text width=21mm] (probe) {\textbf{Feature fusion}\\{\scriptsize linear probe}};
\node[outp, right=7mm of probe] (off) {$\hat{y}$};
\draw[ed] (exp.east) to[out=35,in=180] node[above, pos=0.55, font=\scriptsize] {raw features} (concat);
\draw[ed] (concat) -- (probe);
\draw[ed] (probe) -- (off);

\node[proc, below right=7mm and 14mm of exp, text width=21mm] (cal) {In-fold isotonic\\calibration};
\node[data, right=7mm of cal, text width=16mm] (scores) {Calibrated\\scores $s(x)$};
\draw[ed] (exp.east) to[out=-35,in=180] node[below, pos=0.55, font=\scriptsize] {scalar scores} (cal);
\draw[ed] (cal) -- (scores);
\node[route, above right=3mm and 12mm of scores, text width=25mm] (router) {\textbf{Per-video router}\\{\scriptsize soft weights $w(x)$}};
\node[glob, below right=3mm and 12mm of scores, text width=25mm] (invb) {\textbf{Global \invb}\\{\scriptsize fixed weights}};
\node[outp, right=7mm of router] (oroute) {$\hat{y}$};
\node[outp, right=7mm of invb] (oinvb) {$\hat{y}$};
\draw[ed] (scores) to[out=40,in=180] (router);
\draw[ed] (scores) to[out=-40,in=180] (invb);
\draw[ed] (router) -- (oroute);
\draw[ed] (invb) -- (oinvb);

\begin{scope}[on background layer]
\node[draw=gray!45, dashed, rounded corners=3pt, fit=(scores)(router)(invb), inner sep=5mm, label={[font=\scriptsize\itshape, gray!55!black]below:within the experts' score interval}] {};
\end{scope}
\end{tikzpicture}
\caption{\textbf{The three combiner families.} All read the same four frozen experts. \emph{Top, red:} feature fusion is a linear probe over the \emph{concatenated raw features}; it lies \emph{outside} the experts' score space, so \cref{thm:corner} does not bound it, and we instead audit it empirically (it does not robustly improve on \invb{} under honest \lofo{} folds; its apparent advantage is a fold-granularity artifact, \cref{sec:06_analysis}). \emph{Bottom, dashed box:} the score-space combiners operate on the in-fold-calibrated expert scores $s(x)$, so every routed score lies inside the interval. \emph{Orange:} the per-video router emits soft weights $w(x)$ and is a convex mixture of the scores, so it cannot exceed the hard-corner ceiling (\cref{thm:corner}) and empirically does not beat global fusion. \emph{Green:} the global \invb{} combiner is outperformed on \auc{} by no router we test and also improves the tail; it is our recommended recipe.}
\label{fig:supp_pipelines}
\end{figure*}
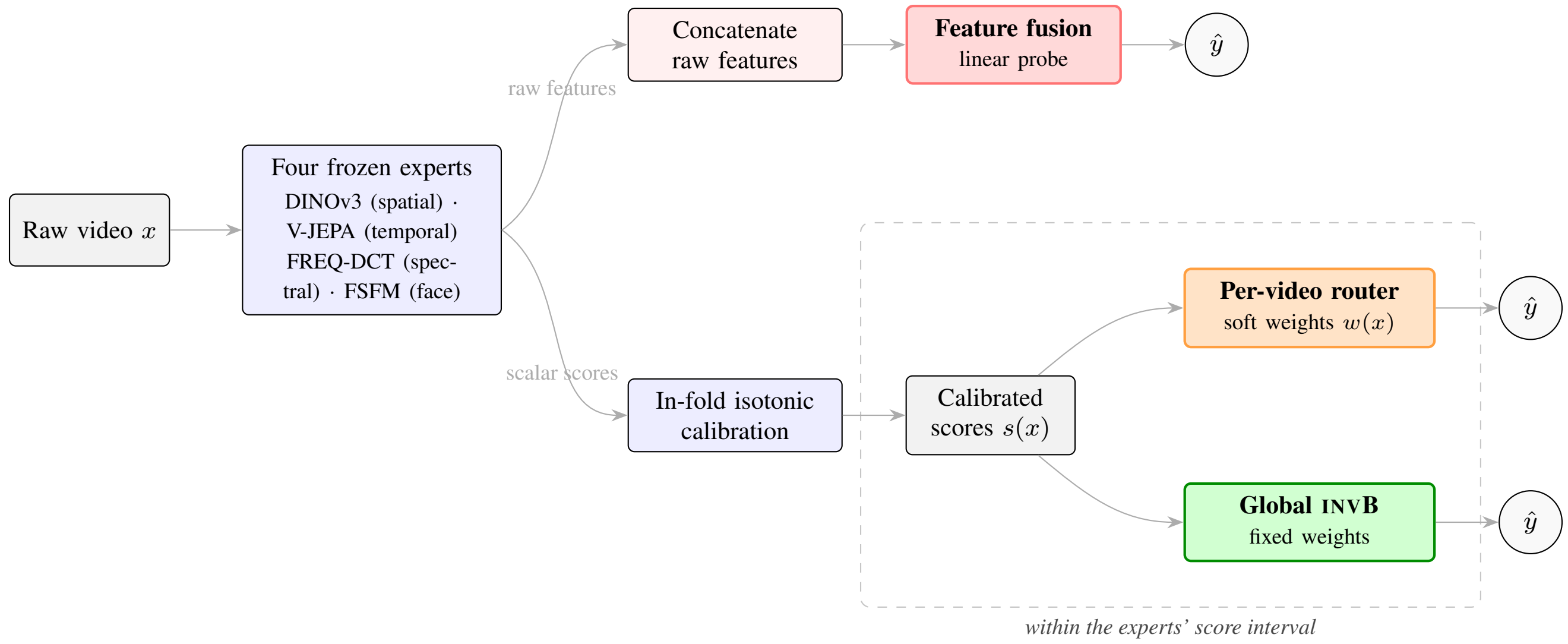

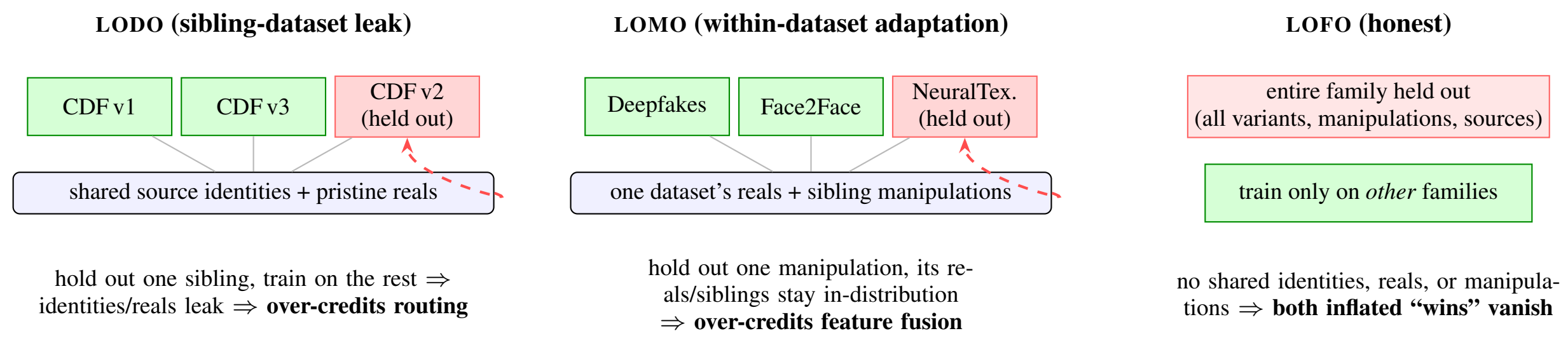
\begin{figure*}[t]
\centering
\begin{tikzpicture}[
  font=\footnotesize,
  >={Stealth[length=1.6mm]},
  cell/.style={draw, minimum width=15mm, minimum height=6mm, align=center, inner sep=2pt, font=\scriptsize},
  train/.style={cell, fill=green!16, draw=green!55!black},
  test/.style={cell, fill=red!16, draw=red!60},
  pool/.style={draw, rounded corners=2pt, fill=blue!6, align=center, inner sep=3pt, font=\scriptsize, minimum width=50mm},
  ttl/.style={font=\footnotesize\bfseries, align=center},
  leak/.style={draw, ->, red!70, thick, dashed},
]

\begin{scope}
  \node[ttl] (t1) at (0,1.55) {\lodo{} (sibling-dataset leak)};
  \node[train] (a1) at (-1.6,0.7) {CDF\,v1};
  \node[train] (a2) at (0,0.7)    {CDF\,v3};
  \node[test]  (a3) at (1.6,0.7)  {CDF\,v2\\(held out)};
  \node[pool]  (p1) at (0,-0.2) {shared source identities + pristine reals};
  \draw[gray!55] (a1) -- (p1); \draw[gray!55] (a2) -- (p1); \draw[gray!55] (a3) -- (p1);
  \draw[leak] (p1.east) to[out=-20,in=-90] (a3.south);
  \node[font=\scriptsize, text width=46mm, align=center] at (0,-1.25)
    {hold out one sibling, train on the rest $\Rightarrow$ identities/reals leak $\Rightarrow$ \textbf{over-credits routing}};
\end{scope}

\begin{scope}[xshift=58mm]
  \node[ttl] (t2) at (0,1.55) {\lomo{} (within-dataset adaptation)};
  \node[train] (b1) at (-1.6,0.7) {Deepfakes};
  \node[train] (b2) at (0,0.7)    {Face2Face};
  \node[test]  (b3) at (1.6,0.7)  {NeuralTex.\\(held out)};
  \node[pool]  (p2) at (0,-0.2) {one dataset's reals + sibling manipulations};
  \draw[gray!55] (b1) -- (p2); \draw[gray!55] (b2) -- (p2); \draw[gray!55] (b3) -- (p2);
  \draw[leak] (p2.east) to[out=-20,in=-90] (b3.south);
  \node[font=\scriptsize, text width=46mm, align=center] at (0,-1.25)
    {hold out one manipulation, its reals/siblings stay in-distribution $\Rightarrow$ \textbf{over-credits feature fusion}};
\end{scope}

\begin{scope}[xshift=116mm]
  \node[ttl] (t3) at (0,1.55) {\lofo{} (honest)};
  \node[test, minimum width=34mm, fill=red!10] (c1) at (0,0.7) {entire family held out\\(all variants, manipulations, sources)};
  \node[train, minimum width=34mm] (c2) at (0,-0.2) {train only on \emph{other} families};
  \node[font=\scriptsize, text width=40mm, align=center] at (0,-1.25)
    {no shared identities, reals, or manipulations $\Rightarrow$ \textbf{both inflated ``wins'' vanish}};
\end{scope}
\end{tikzpicture}
\caption{\textbf{Two evaluation-leakage mechanisms, and how \lofo{} removes both.} Green $=$ in training, red $=$ held out for test. \emph{Left (\lodo):} sibling datasets in a family (e.g.\ the Celeb-DF versions) share source identities and pristine reals, so holding out one sibling while training on the others leaks low-level statistics into the ``unseen'' test set; this inflates the learned \emph{router}. \emph{Middle (\lomo):} holding out one manipulation method leaves that dataset's reals and sibling manipulations in-distribution, letting a feature probe adapt within the dataset; this inflates \emph{feature fusion}. \emph{Right (\lofo):} holding out the whole dataset family at once removes shared identities, reals, and manipulations, and both apparent advantages collapse to noise (\cref{sec:06_analysis}). The fourth protocol, \emph{crossfit} (not drawn), uses random stratified folds, so same-dataset videos sit in train and test simultaneously; it is leakier still, and the three dataset-axis protocols order, in point estimate, into the monotone ladder crossfit $>$ \lodo{} $>$ \lofo{} of \cref{tab:ladder}.}
\label{fig:supp_leakage}
\end{figure*}

\begin{figure*}[t]
\centering
\begin{tikzpicture}[
  font=\footnotesize,
  >={Stealth[length=1.8mm]},
  base/.style={draw, rounded corners=2pt, align=center, inner sep=3pt, minimum height=9mm},
  data/.style={base, fill=gray!10},
  proc/.style={base, fill=blue!7},
  trn/.style={base, fill=green!16, draw=green!55!black},
  tst/.style={base, fill=red!16, draw=red!60, postaction={pattern=north east lines, pattern color=red!70}},
  outp/.style={base, fill=gray!10, draw=black!70, thick},
  ed/.style={draw, ->, gray!65},
  fixed/.style={draw, ->, dashed, green!45!black},
  lane/.style={font=\scriptsize\bfseries, gray!60!black},
]
\begin{scope}[on background layer]
  \fill[gray!4, rounded corners=3pt] (-1mm,-9.5mm) rectangle (160mm,16mm);
  \fill[blue!3, rounded corners=3pt] (-1mm,-48mm) rectangle (160mm,-16.5mm);
\end{scope}
\node[lane, anchor=north west] at (0,15.5mm) {(a) Scoring: once per expert};
\node[data, text width=19mm] (vid) at (11mm,0) {Video $x$\\{\scriptsize $32$ sampled\\frames}};
\node[proc, text width=22mm, right=6mm of vid] (exp) {Expert $k$\\{\scriptsize trained on FF++ c23, then fixed}};
\node[data, text width=26mm, right=6mm of exp] (pf) {per-frame (or per-clip)\\fake probabilities};
\node[proc, text width=20mm, right=6mm of pf] (pool) {mean-pool\\{\scriptsize raw score $\tilde s_k(x)$}};
\node[data, text width=33mm, right=6mm of pool] (join) {join the \K experts per video\\{\scriptsize on (dataset, method, video\_id, label)}\\[1pt]one score matrix};
\begin{scope}[on background layer]
  \node[draw=gray!50, dashed, rounded corners=3pt, fit=(exp)(pf)(pool), inner sep=2mm,
        label={[font=\scriptsize\itshape, gray!55!black, inner sep=1pt]above:for each expert $k=1,\dots,\K$}] {};
\end{scope}
\draw[ed] (vid) -- (exp);
\draw[ed] (exp) -- (pf);
\draw[ed] (pf) -- (pool);
\draw[ed] (pool) -- (join);

\node[lane, anchor=north east] at (159mm,-17mm) {(b) Combiner fitting and evaluation: once per fold};
\node[proc, text width=19mm] (split) at (11mm,-34mm) {protocol split\\{\scriptsize \lofo: hold out\\one family}};
\node[trn, text width=20mm, anchor=west] (tr) at (27mm,-27mm) {training families\\{\scriptsize the other $11$}};
\node[tst, text width=20mm, anchor=west] (te) at (27mm,-41mm) {\phantom{held-out}\\\phantom{family}};
\node[fill=red!10, rounded corners=1pt, inner sep=1.5pt, align=center] at (te.center) {held-out\\family};
\node[proc, text width=25mm, right=6mm of tr] (fitcal) {fit one isotonic\\map $g_k$ per expert};
\node[proc, text width=29mm, right=6mm of fitcal] (fitcomb) {fit the combiner\\{\scriptsize \invb{} weights,\\stack, or gate}};
\node[proc, text width=25mm, right=6mm of te] (appcal) {calibrate\\{\scriptsize $s_k(x)=g_k(\tilde s_k(x))$}};
\node[proc, text width=29mm, right=6mm of appcal] (appcomb) {apply the combiner\\{\scriptsize fused score $\hat s(x)$}};
\node[outp, text width=29mm, anchor=west] (out) at ([xshift=7mm]appcomb.east |- split) {repeat for all $12$ families\\{\scriptsize pool the out-of-fold $\hat s(x)$}\\[1pt]family-macro \auc{}, tail\\{\scriptsize family-cluster bootstrap}};
\draw[ed] (join.south) -- ++(0,-4mm) -| (split.north);
\draw[ed] (split.east) -- ++(1.5mm,0) |- (tr.west);
\draw[ed] (split.east) -- ++(1.5mm,0) |- (te.west);
\draw[ed] (tr) -- (fitcal);
\draw[ed] (fitcal) -- (fitcomb);
\draw[ed] (te) -- (appcal);
\draw[ed] (appcal) -- (appcomb);
\draw[fixed] (fitcal) -- node[right, font=\scriptsize] {fixed} (appcal);
\draw[fixed] (fitcomb) -- node[right, font=\scriptsize] {fixed} (appcomb);
\draw[ed] (appcomb.east) -- ++(2mm,0) |- (out.west);
\end{tikzpicture}
\caption{\textbf{From frame-level predictions to the fused score.} \emph{(a) Scoring}, run once per expert, and per seed for every trained expert (the frozen FSFM is scored once): each expert scores $32$ sampled frames of a video, singly or in clips; its per-frame or per-clip fake probabilities are mean-pooled into one raw score $\tilde s_k(x)$, and the \K experts are joined per video into one score matrix ($43{,}452$ videos at $\K{=}4$), the input to every score-space analysis. \emph{(b) Combiner fitting and evaluation}, run once per fold: the protocol splits the matrix (under \lofo{}, by family; green: training families; red, hatched: held out). Everything learned, namely the per-expert isotonic maps $g_k$ and the combiner (the \invb{} weights, the stack's coefficients, or the gate together with its per-video oracle-winner target; mean fusion has no parameters), is fit on the training families only and then applied, fixed, to the held-out family. Repeating this over all twelve families gives one out-of-fold fused score per video, from which family-macro \auc{}, the tail metrics and the family-cluster bootstrap intervals are computed. Crossfit and \lodo{} change only the split; \lomo{} also changes the evaluation grouping, to (dataset, manipulation) units, the conflation that \cref{sec:2x2} separates (\cref{fig:supp_leakage}); the feature probe uses the same folds, on mean-pooled per-frame features in place of scores (\cref{fig:supp_pipelines}).}
\label{fig:supp_flow}
\end{figure*}
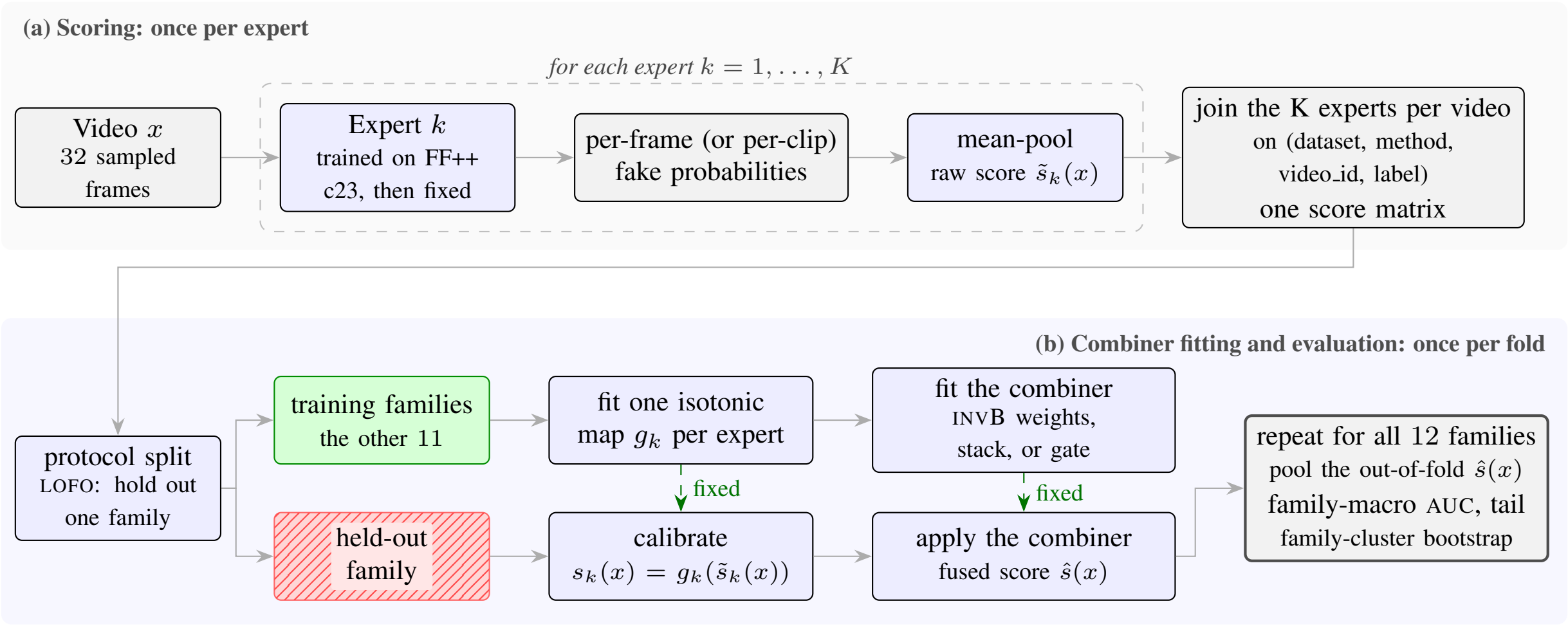

\cref{fig:supp_pipelines} contrasts the three combiner families and shows why the corner theorem (\cref{thm:corner}) governs score-space routing but not feature fusion; \cref{fig:supp_leakage} shows the two evaluation-leakage mechanisms the body audits and how \lofo{} removes both (an expanded version of \cref{fig:protocols}); and \cref{fig:supp_flow} follows a video from frame-level predictions to the fused score, separating what is computed once per expert from what each fold fits on its training families. These figures restate, in schematic form, the setup of \cref{sec:04_methodology} and the leakage analysis of \cref{sec:06_analysis}.

\section{Proof of the Corner Theorem}
\label{supp:proof}
We prove \cref{thm:corner}. Let $K$ experts produce calibrated scores $s(x)=(s_1(x),\dots,s_K(x))\in[0,1]^K$ for video $x$ with label $y\in\{0,1\}$, and let a router output $r(x)=\langle w(x),s(x)\rangle$ for weights $w(x)$ on the simplex $\Delta^{K-1}$.

\textbf{(i) AUC is an ordinal U-statistic.}
For a scoring function $r$, $\mathrm{AUC}(r)=\Pr_{x^+,x^-}\!\big[r(x^+)>r(x^-)\big]+\tfrac12\Pr[r(x^+){=}r(x^-)]$ over an independent positive $x^+$ and negative $x^-$~\cite{menon2016bipartite}.
AUC therefore depends on $r$ only through the \emph{pairwise comparisons} $\mathbf{1}[r(x^+)>r(x^-)]$, i.e.\ it is invariant to any strictly monotone reparametrisation of the score.

\textbf{(ii) Per-pair routing is a vertex LP.}
Fix one pair $(x^+,x^-)$.
The contribution $\mathbf{1}[\langle w(x^+),s(x^+)\rangle>\langle w(x^-),s(x^-)\rangle]$ is, for each argument separately, maximised by a linear objective $\langle w,s\rangle$ over the simplex $\Delta^{K-1}$; a linear program over a simplex attains its optimum at a vertex, i.e.\ at a single expert $e_{k}$.
Routing the positive to its \emph{maximum} expert and the negative to its \emph{minimum} expert (weakly) maximises that pair's indicator, and is a \emph{hard} assignment.

\textbf{(iii) The corner is a label-aware oracle, not the ROC hull.}
Assigning every positive its maximal expert and every negative its minimal expert simultaneously maximises all of the pairwise indicators of step~(ii), so this \emph{label-aware corner} is the \auc-optimal assignment within the per-video intervals and upper-bounds every router.
It uses the label $y$ to pick the favourable extreme per video (an OR over experts on positives, an AND on negatives), so it is strictly stronger than, and must not be identified with, the label-blind upper ROC convex hull of the experts' individual curves~\cite{reid2011information,berta2024classifier}, which fixes one (possibly randomised) operating point independent of the label.
The corner is therefore unrealisable by any label-blind rule; empirically it is the near-perfect ${\approx}\,0.9997$ ceiling, far above any realisable combiner (\cref{tab:supp_theorem}).

\textbf{(iv) Realisable routers and combiners are label-blind; their comparison is empirical.}
A global combiner $g(s)$ and a router are both label-blind functions of $s(x)$.
Neither class contains the other, so the corner of step~(iii) bounds both only as the universal label-aware ceiling and does \emph{not} establish that a router is dominated by a global combiner (or the reverse).
Whether a realisable router matches global fusion is thus an empirical question, answered against routing in \cref{supp:tables} and \cref{sec:05b_main_results}.
We do not claim the ceiling is reachable, or unreachable, in principle for \emph{every} problem; only that it is label-aware, and that here our realisable combiners reach macro AUC $0.949$.

\textbf{(v) Soft routers lie inside the interval.}
A soft router outputs a convex combination of the experts' scores, so $r(x)\in[\min_k s_k(x),\max_k s_k(x)]$ for every $x$: it can only interpolate \emph{inside} that interval.
It cannot exceed the hard-corner assignment of step~(ii) on AUC.
This proves \cref{thm:corner}. \hfill$\square$

\emph{Empirical check (step (v)).}
On all $43{,}452$ videos (\lofo, seed~$42$) the convex combiners (\emph{mean}, \invb, and the per-video gate) lie within $[\min_k s_k(x),\max_k s_k(x)]$ on every video, confirming step~(v); the logistic \emph{stack}, a non-convex global map of the scores rather than a convex mixture, leaves the interval on $34\%$ of videos, placing it in a distinct functional class.

\paragraph{Proper-scoring corollary.}
Steps~(i)--(v) use only the \emph{ordinal} structure of AUC.
Under a strictly proper score such as Brier~\cite{brier1950verification,gneiting2007strictly}, which rewards the \emph{magnitude} of calibrated probabilities, a soft router can strictly beat a hard one by hedging under winner-uncertainty.
However, a single global recalibration of the fixed combiner, one isotonic map fit out-of-fold~\cite{zadrozny2002transforming,guo2017calibration}, lowers mean fusion's pooled Brier from $0.111$ to $0.092$ (the soft gate is $0.090$; \cref{tab:supp_theorem}), shrinking the gap from $0.021$ to $0.003$, so the proper-scoring advantage is overwhelmingly a \emph{calibration} effect attributable to the recalibrator.

\section{Corner-Theorem Evidence and Full Tables}
\label{supp:tables}

\paragraph{Corner-theorem evidence.}
\cref{tab:supp_theorem} reports the realised numbers behind the $K{=}4$ ceiling, separating two provenances: the hard-corner oracle and the Brier gap are \emph{pooled} over all $43{,}452$ videos (seed~$42$), whereas the winner-predictability and hard-routing numbers are \emph{leave-one-family-out} (seed~$42$).
Pooled, a label-aware hard-corner oracle achieves a macro AUC of $0.9997$ (the unrealisable ceiling).
Out-of-family, the oracle winner is highly predictable (the strongest predictor reaches winner-accuracy $q{=}0.84$ versus a $0.39$ four-way majority baseline), yet \emph{hard-routing that predicted winner} scores only $0.940$, significantly \emph{below} \invb (paired $\Delta{=}-0.79$pp, $95\%$ CI $[-1.1,-0.4]$).
Hard-routing the predicted winner therefore loses to \invb even though the winner is predictable: committing to one expert forfeits the cross-expert agreement \invb keeps, so the shortfall is not a weak-gate artifact.
Pooled under Brier, the soft gate is genuinely better than \emph{raw} mean fusion ($0.090$ vs $0.111$, $-19\%$), but this is the calibration effect of the corollary: one global isotonic pass on mean fusion drops its pooled Brier to $0.092$, within $0.003$ of the gate.

\begin{table}[tbp]
\centering
\footnotesize
\setlength{\tabcolsep}{5pt}
\begin{tabular}{lc}
\toprule
quantity ($K{=}4$) & value \\
\midrule
\multicolumn{2}{@{}l}{\emph{Pooled (all $43{,}452$ videos, seed $42$)}}\\
hard-corner oracle, macro AUC (ceiling)          & $0.9997$ \\
soft gate, Brier ($\downarrow$)                  & $0.090$ \\
mean fusion, Brier: raw $\to$ $+1$ isotonic      & $0.111 \to 0.092$ \\
\midrule
\multicolumn{2}{@{}l}{\emph{Leave-one-family-out (seed $42$)}}\\
winner-predictability $q$ (out-of-family)        & $0.84$ \\
\quad four-way majority baseline                 & $0.39$ \\
hard-route predicted winner, macro AUC           & $0.940$ \\
\quad vs \invb, paired $\Delta$                  & $-0.79\text{pp}^{*}$ \\
\bottomrule
\end{tabular}
\caption{\textbf{Corner-theorem evidence.} The pooled oracle ceiling is near-perfect and the oracle winner is highly predictable out-of-family ($q{=}0.84$), yet hard-routing the predicted winner ($0.940$) loses to the calibrated global combiner \invb ($\Delta{=}-0.79$pp, $^{*}$$95\%$ CI excludes $0$): committing to one expert discards cross-expert agreement, so a weak gate does not explain the shortfall. The Brier gap reflects calibration rather than ranking (proper-scoring corollary, \cref{supp:proof}). Pooled rows are over all videos; the winner/routing rows are \lofo, seed~$42$.}
\label{tab:supp_theorem}
\end{table}

\paragraph{Equivalence and power (gate vs.\ mean fusion).}\label{supp:equivalence}
A non-significant difference is not by itself evidence of equivalence, so for the headline routing tie we add a two-one-sided-test (TOST) on the same paired family-cluster bootstrap (\cref{tab:ladder}).
The paired gate$-$mean difference under \lofo is $-0.09$pp with bootstrap SE $0.14$pp (this paired-delta estimator differs by $0.02$pp from the difference-of-seed-means $-0.07$pp reported in \cref{tab:ladder}); the design's minimum detectable effect is $0.39$pp at $80\%$ power ($\alpha{=}0.05$).
TOST certifies equivalence within $\pm0.5$pp ($p{=}0.002$) and $\pm1.0$pp ($p{<}0.001$), but \emph{not} within $\pm0.24$pp ($p{=}0.15$, below the resolvable effect), so routing and mean fusion are indistinguishable on ranking down to ${\approx}0.4$pp.
For feature fusion$-$\invb under \lofo, the held-out family count is smaller, and the per-seed minimum detectable effect is ${\approx}1.9$--$2.5$pp, too large for any sensible equivalence margin; the $+0.28$pp three-seed mean is significant at no seed, so that verdict rests instead on the fold$\times$eval decomposition (\cref{sec:2x2}).
These are reproduced by the released \texttt{run\_equivalence} script; both bootstraps are exact percentile bootstraps, the probe$-$\invb one over the re-extracted per-frame features (too large to ship in the evaluation-only snapshot, whose extraction recipe regenerates them).

\begin{figure}[tbp]
\centering
\includegraphics[width=\columnwidth]{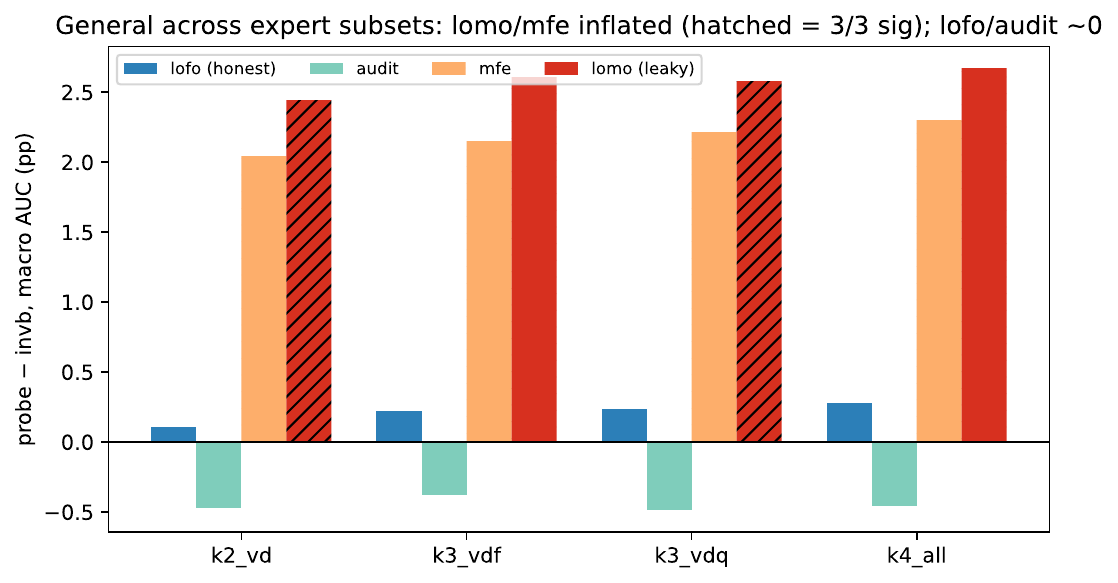}
\caption{\textbf{General across expert subsets.} probe$-$\invb by subset and protocol: manipulation-fold (\lomo) inflated, $3/3$ significant (hatched); honest-fold (\lofo) $\sim 0$. Spans two ViTs, ViT$+$CNN, ViT$+$face; frame-count robust (\cref{sec:generality}).}
\label{fig:supp_bars}
\end{figure}

\paragraph{Full multi-seed decompositions.}
\cref{tab:supp_2x2} expands the fold$\times$eval $2{\times}2$ behind \cref{sec:2x2} and the dataset-axis ladder behind \cref{sec:ladder}, as the grand mean over expert subsets and three seeds $\{42,137,1729\}$ with the count of significant cells (per-subset breakdown in \cref{fig:supp_bars}).
The feature-fusion inflation is a clean main-effect of \emph{fold} granularity: switching the fold from family to manipulation flips probe$-$\invb from ${\sim}0$ to $+2$pp and from $0/12$ to $12/12$ significant cells at \emph{both} eval granularities, while switching the eval granularity barely moves it.
The absence of a robust feature-fusion improvement under honest folds holds under a nested-CV \emph{tuned} probe (\lofo $+0.35$pp, $0/3$ seeds), pre-empting the under-regularisation objection, and the dataset-axis ladder is monotone (crossfit $+3.71$ $[3/3]>$ \lodo $+1.19$ $[3/3]>$ \lofo $+0.28$ $[0/3]$).
The routing ladder of \cref{tab:ladder} mirrors it (gate$-$mean point estimates crossfit $+0.36^{*}\to$ \lodo $+0.16\to$ \lofo $-0.07$), and the tail-robustness gains of \cref{tab:tail} are, by contrast, protocol-stable, the signature that separates a real effect from leakage.

\begin{table}[tbp]
\centering
\footnotesize
\setlength{\tabcolsep}{5pt}
\begin{tabular}{llcc}
\toprule
 & & probe $-$ \invb & \#sig \\
axis & fold (eval) & macro AUC (pp) & cells \\
\midrule
\multirow{4}{*}{fold$\times$eval}
 & honest fold, family eval     & $+0.21$ & $0/12$ \\
 & honest fold, manip.\ eval    & $-0.45$ & $0/12$ \\
 & manip.\ fold, family eval    & $+2.17$ & $12/12$ \\
 & manip.\ fold, manip.\ eval   & $+2.57$ & $12/12$ \\
\midrule
\multirow{3}{*}{\shortstack[l]{dataset-axis\\ladder\\(family eval)}}
 & crossfit fold               & $+3.71$ & $3/3$ \\
 & \lodo fold                  & $+1.19$ & $3/3$ \\
 & \lofo fold                  & $+0.28$ & $0/3$ \\
\bottomrule
\end{tabular}
\caption{\textbf{Feature-fusion leakage decompositions} (grand mean over expert subsets and three seeds; \#sig = significant cells out of $12$ for the $2{\times}2$, out of $3$ for the ladder). Top: the fold$\times$eval $2{\times}2$, a pure main-effect of \emph{fold} granularity. Bottom: the dataset-axis fold ladder at honest (family) eval. No honest-fold cell is significant at any seed; under the pre-audit thirteen-unit grouping, the only significant honest cells were seed-$42$ alone, the single-seed pitfall of \cref{sec:generality}.}
\label{tab:supp_2x2}
\end{table}

\paragraph{A content-conditioned router.}\label{supp:contentrouter}
A natural objection to the routing negative is that our gate reads only score-derived features, so a router that reads the \emph{video} might do better.
\cref{thm:corner} already bounds any such router, since its output is still a convex mixture of the expert scores, and we also test one: the gate input becomes a frozen X-CLIP per-video embedding ($\ell_2$-normalised, dimension $1024$, from a model \emph{outside} the four-expert pool), trained in-fold to predict the per-video oracle winner, in a strongly regularised multinomial-logistic and a gradient-boosted variant, on the same joined videos, \lofo{} folds, and paired family bootstrap as \cref{tab:ladder}.
Soft content-routing ties \invb{} on family-macro \auc{} (logistic $+0.03$pp, $p{=}0.88$; boosted $+0.12$pp, $p{=}0.41$; pooled over three seeds and not significant in any single seed), hard content-routing is significantly worse ($-2.2$pp), and the oracle winner is barely predictable from content at all ($q{\approx}0.43$ from the embedding, versus $0.82$ from the scores and a $0.40$ majority baseline; the $0.82$/$0.40$ pair is this pipeline's three-seed re-fit of the score-derived winner predictor, matched to the embedding gate, consistent with the seed-$42$ ceiling $q{=}0.84$ of \cref{tab:supp_theorem}).
Content conditioning changes the weights but leaves the reachable hull unchanged, so reading the video does not rescue routing and the paper's recommendation stands.
Reproduced by the released \texttt{run\_content\_router}.

\paragraph{Robustness to alternative family definitions.}\label{supp:familyrobust}
The honest \lofo{} verdict (no score-space router we test beats global fusion on \auc, and the only protocol-stable combiner gain is calibration) does not depend on the exact \lofo{} family boundaries.
\cref{tab:supp_familyrobust} re-runs the identical \lofo{} pipeline (the same in-fold isotonic calibration, five combiners, and paired family-cluster bootstrap behind \cref{tab:ladder}) while perturbing only the dataset$\to$family map, in both directions.
Coarser: we fold the one external borderline case, DeeperForensics-1.0, which reuses the FaceForensics++ reals~\cite{jiang2020deeperforensics}, into FFpp (the paper keeps it a separate unit, the leakier choice).
Finer: we split the two within-series merges DFDC/DFDCP and DSv1/DSv2 into singletons, to test those merges; CDF stays merged, as the Celeb-DF versions share the same identities.
Across all three definitions, the routing delta (gate$-$mean) stays non-significant and within $\pm0.16$pp, while the calibration gain (\invb$-$mean) stays positive and significant; the apparent routing gain reappears only as the folds are made finer and leakier still (\lodo, crossfit; \cref{tab:ladder}).
The check is reproduced by the released \texttt{run\_family\_robustness}, which first asserts that the paper grouping reproduces the \cref{tab:ladder} \lofo{} deltas exactly before evaluating the alternatives.

\begin{table}[tbp]
\centering
\footnotesize
\setlength{\tabcolsep}{4pt}
\resizebox{\columnwidth}{!}{\begin{tabular}{lccc}
\toprule
family definition & units & gate$-$mean & \invb$-$mean \\
\midrule
paper (honest \lofo)                    & $12$ & $-0.09$ $[-0.37,+0.17]$ & $+0.27$ $[+0.12,+0.41]^{*}$ \\
\;$+$ DeeperForensics$\to$FFpp (coarser) & $11$ & $-0.09$ $[-0.37,+0.19]$ & $+0.24$ $[+0.10,+0.38]^{*}$ \\
\;split DFDC, DS (finer)                & $14$ & $-0.16$ $[-0.44,+0.11]$ & $+0.23$ $[+0.05,+0.40]^{*}$ \\
\bottomrule
\end{tabular}}
\caption{\textbf{\lofo{} robustness to alternative family definitions.} Paired family-cluster bootstrap macro-\auc{} deltas vs mean fusion (pp, $95\%$ CI; three seeds, FSFM frozen; $2000$ iters), perturbing only the dataset$\to$family map. The routing delta (gate$-$mean) is non-significant under every definition and the calibration gain (\invb$-$mean) is significant under every definition, so the verdict does not hinge on the family boundaries. $^{*}$: $95\%$ CI excludes $0$.}
\label{tab:supp_familyrobust}
\end{table}

\paragraph{Tail robustness under alternative family definitions.}\label{supp:tailfamilyrobust}
The \emph{tail} conclusions of \cref{tab:tail} are likewise grouping-invariant.
\cref{tab:supp_tailfamilyrobust} re-runs the tail pipeline under every family definition, including the original pre-audit grouping with DFDM as a singleton: the \allwrong{} curve falls monotonically, its $K2{-}K4$ drop stays significant, and each fusion-robustness delta (\invb$-$best-single on worst-family \auc, \aurc, and TPR@$1\%$FPR) stays significant under all four definitions, resampling the families of each grouping.
Reproduced by the released \texttt{run\_tail\_family\_robustness}, which first asserts that the original grouping reproduces the stored tail headline before evaluating alternatives.

\begin{table}[tbp]
\centering
\footnotesize
\setlength{\tabcolsep}{3pt}
\resizebox{\columnwidth}{!}{\begin{tabular}{lccccc}
\toprule
family definition & units & \allwrong{} $K2{-}K4$ & worst-family & \aurc{} & TPR@$1\%$ \\
\midrule
paper (DFDM$\to$CDF)          & $12$ & $+4.7^{*}$ & $+4.2^{*}$ & $+2.0^{*}$ & $+21.2^{*}$ \\
\;DFDM singleton (original)    & $13$ & $+4.8^{*}$ & $+4.6^{*}$ & $+2.0^{*}$ & $+20.8^{*}$ \\
\;$+$ DeFo$\to$FFpp (coarser)  & $11$ & $+4.7^{*}$ & $+4.2^{*}$ & $+2.0^{*}$ & $+21.2^{*}$ \\
\;split DFDC, DS (finer)       & $14$ & $+4.8^{*}$ & $+5.2^{*}$ & $+2.0^{*}$ & $+21.6^{*}$ \\
\bottomrule
\end{tabular}}
\caption{\textbf{Tail robustness to alternative family definitions} (honest \lofo; pp; \invb$-$best-single except the \allwrong{} diversity drop and \aurc, reported as its reduction best-single$-$\invb{} so that positive is better; paired family-cluster bootstrap over each grouping's families, three seeds, $2000$ iters). Every tail conclusion survives every definition. $^{*}$: $95\%$ CI excludes $0$.}
\label{tab:supp_tailfamilyrobust}
\end{table}

\paragraph{Excluding FaceForensics++ from evaluation.}\label{supp:ffppexcluded}
Every expert is trained on FF++ c23 (\cref{sec:05a_setup}), and FF++ is also one of the twelve evaluation families, where its videos are in-domain for the experts.
\cref{tab:supp_ffppexcluded} re-runs the evaluation behind \cref{tab:main,tab:ladder,tab:tail} with every FF++ video removed from the pool before any fold is built, so FF++ is absent from both combiner fitting and evaluation; eleven families remain, and the grouping, seeds, experts, metrics, and bootstrap settings are unchanged (no retraining).
We fixed and committed the readout rule before running the check: the conclusions survive if, under \lofo, \invb$-$mean is positive with a CI excluding $0$ and gate$-$mean is not significantly positive, and if the crossfit gate$-$mean point estimate stays above the \lofo{} one.
All three hold: \invb$-$mean is $+0.27$pp $[+0.12,+0.43]$ (with FF++: $+0.27$ $[+0.12,+0.41]$), gate$-$mean is $-0.11$pp $[-0.43,+0.18]$, and the routing ladder runs crossfit $+0.34^{*}\to$ \lodo{} $+0.12\to$ \lofo{} $-0.11$.
Every tail claim keeps its sign and stays significant, and the \allwrong{} curve for $\K{=}1{\to}4$ becomes $15.7{\to}7.3{\to}4.9{\to}2.5\%$.
Absolute levels drop slightly, as FF++ is the easiest family to score: macro \auc{} falls by $0.4$--$0.5$pp for best-single, mean, \invb, and gate, and their ordering is unchanged on every metric of \cref{tab:main}.
Reproduced by the released \texttt{run\_operating\_points}, \texttt{run\_leakage\_table}, and \texttt{run\_tail\_robustness} with \texttt{-{}-exclude-family FFpp}.

\begin{table}[tbp]
\centering
\footnotesize
\setlength{\tabcolsep}{4pt}
\begin{tabular}{lcc}
\toprule
 & $12$ families & $11$ families \\
quantity & (paper) & (no FF++) \\
\midrule
\multicolumn{3}{@{}l}{\emph{Macro} \auc{} \emph{deltas} (pp)}\\
\lofo{} \invb$-$mean                     & $+0.27^{*}$ & $+0.27^{*}$ \\
\lofo{} gate$-$mean                      & $-0.07$     & $-0.11$ \\
\lodo{} gate$-$mean                      & $+0.16$     & $+0.12$ \\
crossfit gate$-$mean                     & $+0.36^{*}$ & $+0.34^{*}$ \\
\midrule
\multicolumn{3}{@{}l}{\emph{Tail} (\lofo; pp)}\\
\allwrong{} $K2{-}K4$                    & $+4.7^{*}$  & $+4.9^{*}$ \\
worst-family \auc{} (\invb$-$best)       & $+4.2^{*}$  & $+4.1^{*}$ \\
\aurc{} (\invb$-$best; $\downarrow$)     & $-2.0^{*}$  & $-2.1^{*}$ \\
TPR@$1\%$FPR (\invb$-$best)              & $+21.2^{*}$ & $+22.5^{*}$ \\
\midrule
\multicolumn{3}{@{}l}{\emph{Levels} (\lofo, \invb)}\\
macro \auc{}                             & $0.949$     & $0.946$ \\
TPR@$1\%$FPR                             & $0.648$     & $0.619$ \\
\bottomrule
\end{tabular}
\caption{\textbf{FF++-excluded check} (three seeds; $^{*}$: $95\%$ paired family-cluster CI excludes $0$). The $12$-family column repeats \cref{tab:main,tab:ladder,tab:tail}. Removing FF++ lowers the absolute levels slightly but changes no sign, no significance verdict, and no ordering of best-single, mean, \invb, and gate; the readout rule, fixed and committed before the run, is met.}
\label{tab:supp_ffppexcluded}
\end{table}

\paragraph{The five-expert tail extension.}\label{supp:tailk5}
A matched X-CLIP~\cite{ni2022expanding} fifth expert, retrained and scored at all three seeds $\{42,137,1729\}$, \emph{stress-tests} the diversity-law prediction (\cref{sec:tail}) and bears it out, without disturbing the four-expert curve: the \allwrong rate falls $K2{\to}K5$ by $+5.3$pp $[+3.5,+6.7]$ (versus $+4.7$pp at $K{=}4$) and the full five-expert tail drops from $2.3\%$ to $1.8\%$, while the fusion-robustness legs stay significant (worst-family $+3.8^{*}$, \aurc $-1.8^{*}$, TPR@$1\%$ $+20.1^{*}$pp).
The combiner verdict is unchanged, indeed sharpened: at $\K{=}5$ the honest-\lofo gate is significantly worse than mean fusion ($-0.40$pp $[-0.77,-0.03]$, below mean fusion in all three seeds), the fifth expert giving the router more to overfit, while \invb$-$mean stays significant in both crossfit and \lofo ($+0.20$pp).
Under crossfit the soft-gate Brier improves from $0.0897$ at $K{=}4$ to $0.0888$ at $K{=}5$ against a mean-fusion $0.1131$ (a $21.4\%$ reduction, $3/3$ seeds); this growth is crossfit-specific, and under honest \lofo the calibration win is protocol-stable but does not grow with $\K$ (isotonic Brier reduction $15.7\%$ at $K{=}4$ versus $14.5\%$ at $K{=}5$).
All five-expert numbers are regenerated by the released \texttt{run\_tail\_robustness} and \texttt{run\_multiseed\_summary} with \texttt{-{}-experts vjepa dinov3 fsfm freq xclip}.

\paragraph{Scope of the generality evidence.}\label{supp:generalityscope}
Every analysis in the paper takes the calibrated per-video score matrix of the fixed experts as its input: the leakage audit, the corner-theorem evidence, and the tail metrics are computed from that matrix and the folds, so each claim is relative to the combiner on this matrix and none hinges on reproducing a given expert's training.
Whether the same patterns hold for another, comparably diverse pool is an empirical question that we test only in part, and only within this $22$-dataset testbed.
The evidence spans five architecturally diverse experts, including a video-language model (X-CLIP), three seeds, and two frame counts: both leakage ladders hold at every seed, the feature-fusion artifact also holds across the four expert subsets (\cref{fig:supp_bars}) and in a $64$-frame re-score (\cref{sec:generality}), and with the fifth expert the diversity law and the fusion-robustness legs stay significant while the gate becomes significantly worse than mean fusion (\cref{supp:tailk5}).
We do not claim these patterns for arbitrary expert pools.

\section{Combiner Implementation Details}
\label{supp:impl}
Every combiner is fit strictly out-of-fold: within each training fold, we first fit one isotonic calibrator per expert (monotone regression with out-of-range clipping~\cite{zadrozny2002transforming}) on that fold's videos only, and every combiner below then reads only the calibrated training-fold scores.

\paragraph{Gate (per-video router).}
The gate is a gradient-boosted classifier (histogram-based gradient boosting, library defaults, seeded) trained per fold to predict the per-video \emph{oracle winner}: the index of the expert with the lowest per-video binary cross-entropy on its calibrated score; ties are resolved to the lowest index.
Its input is a $14$-dimensional score-derived vector per video at $\K{=}4$: the four calibrated scores $s_k$, the four confidences $|s_k{-}0.5|$, and the six pairwise disagreements $|s_i{-}s_j|$.
At test time the classifier's soft class-probabilities over the $\K$ experts act as routing weights, and the routed score is $\langle w(x), s(x)\rangle$; hard routing instead commits to $\arg\max_k w_k(x)$.
If a training fold contains a single winner class, the gate degenerates to always choosing that expert.

\paragraph{Global combiners.}
The stack is an $\ell_2$-regularised logistic regression over the $\K$ calibrated scores (up to $2000$ iterations); \invb weights expert $k$ by the inverse of its training-fold Brier score, normalised over experts; mean fusion is parameter-free.

\paragraph{Feature probe.}
The feature-fusion probe consumes per-video features (per-frame backbone features mean-pooled over frames), applies per-expert $\ell_2$ normalisation, concatenates the expert blocks, standardises within the training fold, and fits an $\ell_2$-regularised logistic probe.
The inverse-regularisation strength is selected in the strong-regularisation regime on training folds only; the nested-CV variant of \cref{supp:tables} selects it by an inner grouped cross-validation over a log grid, so no probe hyperparameter sees the held-out family.
Under-regularisation is itself an out-of-distribution trap: at the library default, the probe drops significantly below mean fusion, which is why the audit reports the tuned (nested) probe alongside the fixed one.

\paragraph{Computational cost.}\label{supp:efficiency}
\cref{tab:supp_efficiency} reports the per-video inference cost of every expert as scored (32 sampled frames per video): parameters, multiply-accumulates from a per-video forward FLOP count, and wall clock at the scoring batch size on one RTX~4090 under the scoring path's fp16 autocast (forward pass only; decode and face cropping excluded, and X-CLIP encodes 16 of the 32 frames because it subsamples each clip to its pretrained 8-frame context).
Fusion's cost is the $\K$ backbone passes themselves: the four-expert suite totals $5{,}889$ GMACs and $202$\,ms per video, $2.7\times$ the compute ($2.2\times$ the wall clock) of the strongest single expert, DINOv3.
Score-space routing saves none of it, because the gate reads all $\K$ calibrated scores, so every expert must run on every video; the gate itself is a $14$-input classifier costing ${\sim}1.9\,\mu$s per video on CPU, and \invb (four isotonic maps plus a fixed weighted sum) costs $0.15\,\mu$s, six orders of magnitude below one DINOv3 pass, so the choice \emph{among} score-space combiners is computationally free. The only real compute lever is $\K$.
The content router of \cref{supp:contentrouter} adds a fifth backbone (X-CLIP, $+22\%$ GMACs, $+47$\,ms) on top of, not instead of, the four expert passes.

\begin{table}[tbp]
\centering
\footnotesize
\setlength{\tabcolsep}{3.5pt}
\resizebox{\columnwidth}{!}{\begin{tabular}{lrrrrr}
\toprule
 & params & train. & GMACs & ms/video & peak \\
expert & (M) & (M) & /video & @4090 & (GB) \\
\midrule
DINOv3 ViT-L/16 (LoRA)        & $333.5$ & $30.4$  & $2193$ & $93.9$ & $6.1$ \\
V-JEPA 2.1 ViT-L (LoRA)       & $305.5$ & $1.6$   & $1662$ & $67.9$ & $5.1$ \\
FREQ-DCT CNN                  & $1.0$   & $1.0$   & $64$   & $5.0$  & $3.8$ \\
FSFM ViT-L/16 (frozen)        & $303.3$ & $0.002$ & $1970$ & $35.7$ & $2.7$ \\
\midrule
$\K{=}4$ (fusion or routing)  & $943.4$  & $33.0$ & $5889$ & $202.4$ & $6.1$ \\
\;$+$ X-CLIP ViT-L/14 ($\K{=}5$) & $1522.2$ & $36.2$ & $7196$ & $249.8$ & $6.1$ \\
\bottomrule
\end{tabular}}
\caption{\textbf{Per-video inference cost as scored} ($32$ sampled frames per video; frame experts at $224^2$, clip experts $2{\times}16$ frames at $256^2$). Trainable = LoRA adapters $+$ head. GMACs from a per-video forward FLOP count (\texttt{torch.utils.flop\_counter}, GMACs $=$ FLOPs$/2$); ms/video at each expert's scoring batch size under fp16 autocast, warmup $10$ $+$ $50$ timed iterations, synchronised; forward pass only (decode and face cropping excluded). Experts run sequentially, so the suite peak is the max, not the sum, and suite ms/video is the serial single-GPU sum. Score-space combiners add ${\le}2\,\mu$s per video on CPU (\invb $0.15\,\mu$s, stack $0.01\,\mu$s, gate $1.9\,\mu$s), so routing and fusion cost the same $\K$ backbone passes.}
\label{tab:supp_efficiency}
\end{table}

\section{Reproducibility}
\label{supp:repro}
{\raggedright We release the code and an evaluation-only snapshot at \coderepo{}.\par}
\noindent The snapshot regenerates every number, table, and figure from precomputed per-expert scores: no GPU or model weights are required, since all analysis is post-processing on per-video predictions.
Scores are joined on \texttt{(dataset, method, video\_id, label)} so paired-identity datasets retain both classes and every manipulation of a source id stays a distinct video; every per-expert isotonic calibrator is fit inside training folds only, and no test score is used to fit any calibrator, combiner, or router.
All combiners, protocols, metrics, and figures are produced by the released code (\texttt{run\_operating\_points}, \texttt{run\_feature\_probe} with the fold/eval switches, \texttt{run\_tail\_robustness} and \texttt{run\_multiseed\_summary} with the \texttt{-{}-experts} switch, \texttt{run\_generality\_summary/figures}, \texttt{run\_equivalence}, \texttt{run\_family\_robustness}, and \texttt{io\_join}); seeds, folds, and the join are fixed and documented, and a single entry point regenerates the operating points, the corner-theorem evidence (\cref{tab:supp_theorem}), the equivalence tests (\cref{supp:equivalence}), the leakage decompositions (\cref{tab:supp_2x2}), and the tail ladder.
 
\end{document}